\documentclass[letterpaper, 10 pt, journal, twoside]{IEEEtran}
\IEEEoverridecommandlockouts                              % This command is only needed if 
\usepackage{00_ps}

\preprinttrue% Either \preprinttrue or \preprintfalse
\publishedfalse% Either \publishedtrue or \publishedfalse

\glsdisablehyper % disable hyperlinks for glossary entries
\DeclareAcronym{cav}{
    short = CAV,
    long  = Connected and Automated Vehicle,
}
\DeclareAcronym{cbf}{
    short = CBF,
    long  = Control Barrier Function,
}
\DeclareAcronym{clf}{
    short = CLF,
    long  = Control Lyapunov Function,
    short-indefinite = a,
}
\DeclareAcronym{cpmlab}{
    short = CPM Lab,
    long  = Cyber-Physical Mobility Lab
}
\DeclareAcronym{mappo}{
    short = MAPPO,
    long  = Multi-Agent \acl{ppo},
    short-indefinite = an,
}
\DeclareAcronym{maddpg}{
    short = MADDPG,
    long  = Multi-Agent Deep Deterministic Policy Gradient,
    short-indefinite = an,
}
\DeclareAcronym{mas}{
    short = MAS,
    long  = Multi-Agent System,
    short-indefinite = an,
}
\DeclareAcronym{mdp}{
    short = MDP,
    long  = Markov Decision Process,
    short-indefinite = an,
}
\DeclareAcronym{mg}{
    short = MG,
    long  = Markov Game,
    short-indefinite = an,
}
\DeclareAcronym{mpc}{
    short = MPC,
    long  = Model Predictive Control,
    short-indefinite = an,
}
\DeclareAcronym{marl}{
    short = MARL,
    long  = Multi-Agent \acl{rl},
    short-indefinite = an,
}
\DeclareAcronym{ocp}{
    short = OCP,
    long  = Optimal Control Problem,
    short-indefinite = an,
    long-indefinite = an,
}
\DeclareAcronym{per}{
    short = PER,
    long  = Prioritized Experience Replay
}
\DeclareAcronym{pet}{
    short = PET,
    long  = Post-Encroachment Time,
}
\DeclareAcronym{pomdp}{
    short = POMDP,
    long  = Partially Observable \acl{mdp}
}
\DeclareAcronym{pomg}{
    short = POMG,
    long  = Partially Observable \acl{mg}
}
\DeclareAcronym{ppo}{
    short = PPO,
    long  = Proximal Policy Optimization
}
\DeclareAcronym{rl}{
    short = RL,
    long  = Reinforcement Learning,
    short-indefinite = a,
}
\DeclareAcronym{sat}{
    short = SAT,
    long  = Separating Axis Theorem,
}
\DeclareAcronym{sdf}{
    short = SDF,
    long  = Signed Distance Function,
    short-indefinite = an,
}
\providecommand{\doi}{????/placeholder-doi} % <--- Change here if \publishedtrue

\makeatletter
\def\ps@IEEEtitlepagestyle{%
    \def\@oddfoot{\mycopyrightnotice}%
    \def\@evenfoot{}%
}
\makeatother

\ifpreprint
    \ifpublished
        \def\mycopyrightnotice{%
            \begin{minipage}{\textwidth}
                \centering \scriptsize
                \href{https://doi.org/\doi}{DOI: \doi} \copyright \the\year{} \copyrightOwnerFull{}. 
                Personal use of this material is permitted. Permission from \copyrightOwnerShort{} must be obtained for all other uses, in any current or future media, including reprinting/republishing this material for advertising or promotional purposes, creating new collective works, for resale or redistribution to servers or lists, or reuse of any copyrighted component of this work in other works.\hfill
            \end{minipage}
            \gdef\mycopyrightnotice{}% just in case
        }
    \else % preprint, submitted
        \def\mycopyrightnotice{%
            \begin{minipage}{\textwidth}
                \centering \scriptsize
                This work has been submitted to the \copyrightOwnerFull{} for possible publication. Copyright may be transferred without notice, after which this version may no longer be accessible.
            \end{minipage}
            \gdef\mycopyrightnotice{}% just in case
        }
    \fi
\else
    \def\mycopyrightnotice{}
\fi

\def\BibTeX{{\rm B\kern-.05em{\sc i\kern-.025em b}\kern-.08em
  T\kern-.1667em\lower.7ex\hbox{E}\kern-.125emX}}

\begin{document}

\title{
    Safeguarding Multi-Agent Reinforcement Learning for Connected and Automated Vehicles with Recursively Feasible Control Barrier Functions
}

\author{
    Jianye Xu$^{1}$\,\orcidlink{0009-0001-0150-2147},~\IEEEmembership{Student~Member,~IEEE},
    Yousef Sayed$^{2}$\,\orcidlink{0009-0000-0487-3640},
    Bassam Alrifaee$^{2}$\,\orcidlink{0000-0002-5982-021X},~\IEEEmembership{Senior~Member,~IEEE}% <-this % stops a space
    \thanks{$^{1}$Department of Computer Science, RWTH Aachen University, Germany, \texttt{xu@embedded.rwth-aachen.de}}
    \thanks{$^{2}$Department of Aerospace Engineering, University of the Bundeswehr Munich, Germany, \texttt{yousef.sayed@tum.de}, \texttt{bassam.alrifaee@unibw.de}}
}
    \maketitle
\thispagestyle{IEEEtitlepagestyle}
\begin{abstract}
Multi-Agent Reinforcement Learning (MARL) shows promise in coordinating connected and automated vehicles but offers no safety guarantee. Although Control Barrier Functions (CBFs) can add this guarantee, their optimization problem can become infeasible in practice due to control bounds. Moreover, they often overapproximate vehicle and road geometry, which makes them conservative. This paper proposes the Backup-Horizon CBF (BH-CBF), which evaluates safety along a horizon until all vehicles stop. BH-CBF is recursively feasible under control bounds, and it considers the exact vehicle and road geometry. Based on BH-CBF, we derive a fully decentralized safety filter that only needs a single broadcast of vehicle states and routes per step and stays real-time in dense traffic. We use BH-CBF to safeguard a MARL policy over its life cycle: as a reward during pre-training, as a run-time filter at deployment, and as a filter in the training loop that keeps fine-tuning on unseen maps collision-free. In simulations on real-world maps of a roundabout and an intersection, BH-CBF avoids all collisions of unsafe MARL policies and keeps its safety guarantee at every step. Existing filters either collide, lose the guarantee, or fall back to full braking far more often. In recorded roundabout traffic with 1264 human-driven vehicles over 115 km, our safeguarded policies shorten the median travel time by 14\% compared with the human drivers, without any collision.
\end{abstract}

\begin{IEEEkeywords}
Connected and autonomous vehicles, control barrier functions, methods for safety, multi-agent reinforcement learning, optimization and control
\end{IEEEkeywords}

\acresetall
\section{Introduction}\label{sec:intro}

Multi-Agent Reinforcement Learning (MARL) policies have become a promising approach to coordinating Connected and Automated Vehicles (CAVs) in interactive traffic such as merges, lane changes, intersections, and roundabouts \cite{dinneweth2022multiagent,chen2023deep,han2024multiagent,xu2024sigmarl}. Such policies are fast at run time and scale with the number of vehicles, and they can learn cooperative behavior from interaction instead of relying on hand-designed coordination rules \cite{dinneweth2022multiagent}. However, a learned policy offers no safety guarantee since it is trained to maximize a reward that scores its behavior. Collision penalties reduce collisions but do not exclude them, and the behavior of the policy degrades in scenarios that differ from training, such as unseen road layouts \cite{kirk2023survey} or the transfer from simulation to real vehicles \cite{beerwerth2026zeroshot}. Therefore, safety-critical deployment requires a safeguard for the learned policy that does not depend on how well the policy generalizes.

\begin{figure}[t]
\centering
\begin{subfigure}{0.49\columnwidth}
\centering
\includegraphics[width=\linewidth]{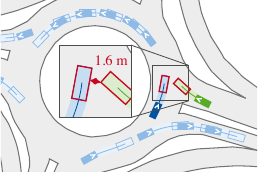}
\caption{A roundabout.}
\label{fig:headline-round}
\end{subfigure}\hfill
\begin{subfigure}{0.49\columnwidth}
\centering
\includegraphics[width=\linewidth]{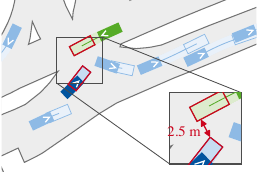}
\caption{An intersection.}
\label{fig:headline-ind}
\end{subfigure}
\caption{An unsafe MARL policy safeguarded by our BH-CBF. Light trails: predictions of braking to a standstill along the route. Red: standstill distance of a vehicle pair.}
\label{fig:headline}
\end{figure}

Safeguards for learned policies can be classified by how they modify the action proposed by the policy \cite{krasowski2023provably}, which we call the \emph{nominal action}. Action masking restricts the action space to verified actions, action replacement substitutes a verified action, as in shields \cite{bastani2021safea}, and action projection moves the action to the closest safe one. We focus on action projection, which keeps interventions small. Control Barrier Function (CBF) safety filters are among the most prominent projection methods, where a Quadratic Program (QP) modifies the nominal action minimally so that a barrier condition holds~\cite{krasowski2023provably}. The barrier condition bounds how fast a safety margin (e.g., the distance to the nearest vehicle) may shrink in the next step. It thereby keeps the system in a safe set of states at all future times, i.e., it renders the safe set forward invariant \cite{ames2017control,ames2019control}. Since CBF filters work with any nominal planner, they suit learned policies \cite{cheng2019endtoend}. Learned graph CBFs even scale to more than a thousand agents, although their learned barriers are not verified \cite{zhang2025gcbf}. For CAVs, MARL policies have been safeguarded by replacing unsafe merging actions with those of a priority-based supervisor \cite{chen2023deep} and by testing discrete lane-change actions for the feasibility of a CBF-QP \cite{han2024multiagent}. However, two shortcomings of existing CBF filters limit their use for CAVs in dense urban traffic.

First, existing CBF filters often overapproximate the vehicle and road geometry. Multi-robot CBFs commonly model agents as disks \cite{wang2017safety}. Lane-keeping CBFs bound the lateral offset of a reference point from the lane center \cite{ames2017control,kalaria2023safety} or shrink the admissible lateral offsets and yaw angles of a rectangle on a straight lane to an inscribed ellipse \cite{voros2025lanekeeping}. Such approximations remove the lateral space a rectangular vehicle has in a narrow lane, which makes the filter conservative, lowers throughput, or even causes deadlocks. Optimization inside the filter can compute exact distances between convex shapes, but these methods solve one centralized QP over all robots \cite{thirugnanam2022dualitybased} or address a single robot \cite{thirugnanam2022safetycritical}. Moreover, the distance is zero for overlapping shapes, so it gives no direction to recover from a collision. The largest gap along the edge normals of two polygons is a closed-form lower bound of their signed distance that is negative when they overlap \cite{han2024efficient}. However, its smoothed version is proven to be a conservative continuous-time CBF only for two single integrators with unbounded inputs \cite{wu2025optimizationfree}.

Second, existing CBF filters lack recursive feasibility, i.e., the guarantee that the filter finds an admissible control that satisfies the barrier condition at every future step. This lack is the most consequential limitation for safeguarding MARL policies. A function whose nonnegative values define the safe set is only a candidate CBF. It is a valid CBF if such a control exists at every safe state, and only then does the filter remain feasible and keep the safe set forward invariant \cite{lindemann2019control,lindemann2019decentralized}. Under control bounds, no general method yields a valid CBF, so candidate CBFs are often used in practice without this guarantee \cite{xiao2022highorder,so2024how}. This is because the barrier condition is myopic: it constrains only the next step, whether a continuous-time condition is enforced at the sampling instants \cite{kalaria2023safety,voros2025lanekeeping,wu2025optimizationfree} or a discrete-time condition is imposed on a one-step prediction \cite{agrawal2017discrete}. Some constraints are affected by the control indirectly. For example, an acceleration command changes the distance between two vehicles only through their speeds. The High-Order CBF handles such constraints by chaining a set of class \(\mathcal K\) functions \cite{xiao2022highorder}, whereas the Truncated Taylor CBF (TTCBF) of our prior work imposes the condition on a truncated Taylor prediction with a single class \(\mathcal K\) function \cite{xu2025highorder,xu2026ttcbf}. In dense traffic, vehicles approach each other quickly when merging or entering a roundabout and come close to each other or to a lane boundary. Then no control within control bounds satisfies the condition, the QP becomes infeasible, and the safety guarantee is lost. CBF-based merging of CAVs \cite{xiao2021bridging} and a CBF shield for MARL-based CAV planning \cite{han2024multiagent} share this limitation: both guarantees assume that the QP is feasible at every step. The shield falls back to an emergency stop without a safety guarantee when the QP is infeasible.

Several families of filters look beyond one step and preserve the ability of a fallback maneuver to keep the system safe \cite{hsu2024safety}. Hamilton–Jacobi reachability analysis yields the least restrictive filter, but its computational cost grows exponentially with the state dimension, so it cannot be applied to many CAVs \cite{hsu2024safety}. Predictive safety filters plan a whole input sequence that ends in a set known to be safe \cite{wabersich2021predictive}. Most methods use a fixed backup policy, which is a fallback maneuver such as emergency braking, and require that the backup keeps the system safe from the state that the applied control reaches. Model predictive shielding \cite{bastani2021safea,zhang2019mamps}, \texttt{gatekeeper} \cite{agrawal2024gatekeeper}, and fail-safe trajectories \cite{pek2021failsafe} use action replacement: they substitute the backup control for an unsafe nominal action, so an intervention can jump from the nominal action to full braking. Backup CBFs impose continuous-time conditions along the backup over a fixed horizon that must end in a known safe set \cite{gurriet2020scalable,chen2021backup}. The smallest safety margin along a fixed policy is itself a discrete-time CBF \cite{yin2025safe}. It has been learned without verification \cite{yin2025safe}, evaluated by finite rollouts in continuous time \cite{knoedler2025safety}, and derived for a single controlled vehicle on a straight road \cite{squires2025barrier}. A related reach-avoid value has been approximated online by differential dynamic programming \cite{rameshkumar2023fast}. For CAVs, these backup-based filters either switch abruptly to the backup, require a fixed horizon ending in a synthesized terminal set, or are limited to one vehicle or to barriers without a guarantee.

Many interacting CAVs further require decentralized computation, because each vehicle must determine its own control from what it observes or receives. The core difficulty is to ensure that the control one vehicle optimizes does not conflict with the controls that the other vehicles optimize at the same time~\cite{alrifaee2016coordinated}. For a CBF, a pair condition couples the controls of two vehicles, so it must be split into one part per vehicle such that the two parts together imply it. Barrier certificates for multiple robots split each pair condition between the two robots in proportion to their acceleration limits \cite{wang2017safety}. In the decentralized backup CBF, each robot covers half of the allowed decrease of a smooth pair barrier \cite{chen2021guaranteed}. This split is exact only in continuous time, where the rate of change of the barrier is the sum of one term per robot. Over a finite sampling step, the motions of the two robots interact, so the two halves no longer imply the pair condition. Other methods split the space instead of the condition and keep the planning problems recursively feasible. They separate the agents by hyperplanes between communicated trajectories \cite{park2022online} or by planes that bisect the braking positions of spherical agents, which needs no communication \cite{schildbach2026contingency}. None of these splits a pair condition so that it remains valid over a finite step for rectangular vehicles whose heading changes.

Safeguarding a MARL policy also takes more than filtering its actions. A filter alone leaves the policy unchanged, so a policy that relies on it keeps proposing unsafe actions. A CBF can also make the policy itself safer. Its condition can serve as a reward that penalizes violations during training, as in our prior work \cite{xu2026safety}. The policy can also be trained with the filter in the loop \cite{cheng2019endtoend,krasowski2023provably}. Both uses inherit the myopia of the one-step condition. The reward penalizes an unsafe action only when a violation is imminent, and an infeasible filter cannot avoid collisions with guarantees.

In summary, a decentralized CBF that is recursively feasible under control bounds and considers the exact vehicle and road geometry is still missing. Nor does any CBF for CAVs look far enough ahead to serve as a reward or as a filter in the training loop of a MARL policy. This paper closes both gaps with the Backup-Horizon CBF (BH-CBF). Our contributions are:

\begin{enumerate}
\def\labelenumi{\arabic{enumi}.}
\item
  We propose BH-CBF, which evaluates the safety margins along a backup policy that brakes every vehicle to a standstill along its route (light trails in \cref{fig:headline}). This horizon makes BH-CBF recursively feasible under control bounds, i.e., a valid CBF, and its margin considers the exact vehicle and road geometry.
\item
  We derive a fully decentralized safety filter from BH-CBF. Each vehicle computes its own control from one broadcast of vehicle states and routes per step, and the computation time stays real-time in dense traffic.
\item
  We show how BH-CBF safeguards a MARL policy over its life cycle: as a reward in pre-training, as a run-time filter at deployment, and as a filter in the training loop that keeps fine-tuning on unseen maps collision-free.
\end{enumerate}
We evaluate BH-CBF on real-world maps of a roundabout and an intersection against state-of-the-art one-step CBF filters, the backup CBF, and model predictive shielding. In recorded roundabout traffic with 1264 human-driven vehicles, we demonstrate that our safeguarded policies shorten the median travel time of the human drivers by \SI{14}{\percent} without any collision. Video demonstrations and the code that reproduces our results are available at \url{https://bassamlab.github.io/BH-CBF}.

This paper extends our prior work. Our filter for polyline lane boundaries \cite{xu2025realtime} covered the vehicle with circles and used a smoothed pseudo-distance, but it is limited to a single vehicle. Our learned pair barrier \cite{xu2025learningbased} approximated the separation of two rectangles by a neural network. BH-CBF uses the exact vehicle geometry instead and handles many vehicles. In \cite{xu2026ttcbf}, we suggested extending TTCBF to a longer look-ahead with the input held constant but left its theory open, which BH-CBF settles with an exact prediction along a backup policy. Our one-step CBF reward \cite{xu2026safety} used the TTCBF condition with circular footprints, and BH-CBF replaces both.

The rest of the paper is organized as follows. \Cref{sec:problem} states the problem. \Cref{sec:bh-cbf} presents BH-CBF with its guarantees, and \cref{sec:safeguard} describes how it safeguards MARL policies. \Cref{sec:evaluation} reports the evaluation, and \cref{sec:conclusion} concludes.

Throughout the paper, vectors, matrices, and vector-valued functions are bold, e.g., \(\bm x\) and \(\bm f\); sets are calligraphic. A superscript \(i\) or \(ij\) refers to vehicle \(i\) or to the pair \((i, j)\); other indices, such as the time step \(k\), are subscripts. A prediction made at step \(k\) for step \(k+t\) carries the subscript \(k+t|k\), e.g., \(\bm x_{k+t|k}\). Joint quantities of all vehicles are written without the vehicle superscript, e.g., \(\bm x = (\bm x^1, \dots, \bm x^n)\). The Euclidean inner product and norm are \(\langle\cdot,\cdot\rangle\) and \(\|\cdot\|\), and \(\lceil\cdot\rceil\) is the ceiling.

\section{Problem Formulation}\label{sec:problem}

We consider \(n\) CAVs with the index set \(\mathcal I = \{1, \dots, n\}\) on a road given by a high-definition map in the OpenStreetMap format \cite{haklay2008openstreetmap}. Each vehicle follows a reference route whose centerline and left and right boundaries are polylines, and the boundaries may be nonconvex and have sharp vertices.

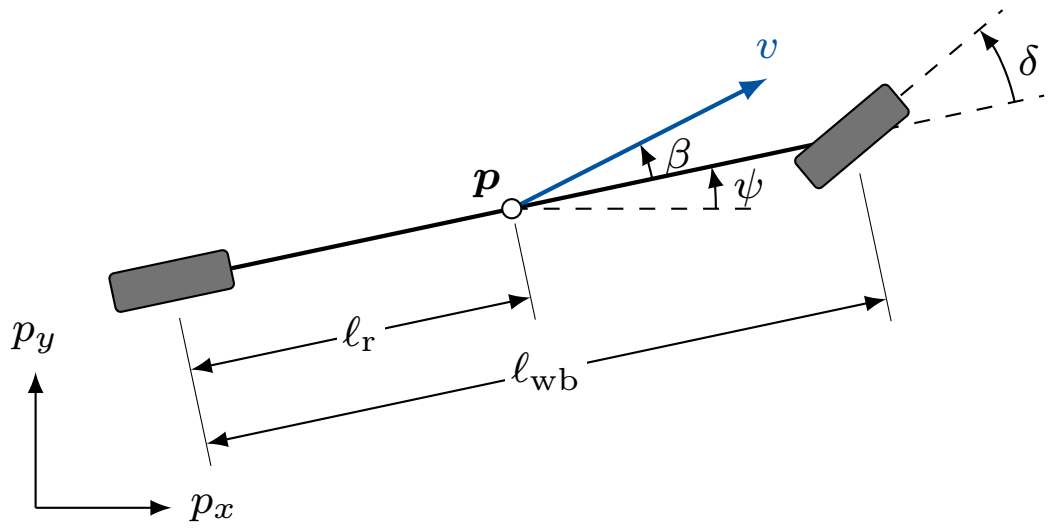
\begin{figure}[t]
\centering
% Kinematic bicycle model; notation as in the text.
% Angles in degrees: heading 12, steering 28, slip atan(tan(28)/2) ~ 15 for l_r = l_wb/2.
\begin{tikzpicture}[
    font=\footnotesize,
    >=latex,
    wheel/.style={draw, fill=black!55, rounded corners=1pt},
    dim/.style={<->, thin},
    ext/.style={very thin},
  ]
  \def\hdg{12}
  \def\steer{28}
  \def\slip{15}
  \def\lwb{4.6}
  \def\lr{2.3}

  % Vehicle frame: rear axle at the origin, x-axis along the vehicle
  \begin{scope}[rotate=\hdg]
    \coordinate (P) at (\lr, 0);
    \draw[thick] (0, 0) -- (\lwb, 0);
    \draw[dashed] (\lwb, 0) -- (\lwb + 1.3, 0);
    \draw[wheel] (-0.4, -0.13) rectangle (0.4, 0.13);
    \begin{scope}[shift={(\lwb, 0)}, rotate=\steer]
      \draw[dashed] (0, 0) -- (1.3, 0);
      \draw[wheel] (-0.4, -0.13) rectangle (0.4, 0.13);
    \end{scope}
    \draw[->] (\lwb + 1.1, 0) arc[start angle=0, end angle=\steer, radius=1.1]
      node[pos=0.55, right] {\(\delta\)};

    % Dimensions
    \draw[ext] (0, -0.25) -- (0, -1.25);
    \draw[ext] (\lr, -0.1) -- (\lr, -0.75);
    \draw[ext] (\lwb, -0.25) -- (\lwb, -1.25);
    \draw[dim] (0, -0.6) -- node[fill=white, inner sep=1pt] {\(\ell_{\mathrm r}\)} (\lr, -0.6);
    \draw[dim] (0, -1.1) -- node[fill=white, inner sep=1pt] {\(\ell_{\mathrm{wb}}\)} (\lwb, -1.1);
  \end{scope}

  % Heading, slip angle, and velocity at the reference point p
  \draw[dashed] (P) -- ++(1.6, 0);
  \draw[->] ([shift={(1.35, 0)}]P) arc[start angle=0, end angle=\hdg, radius=1.35]
    node[pos=0.5, right] {\(\psi\)};
  \draw[->] ([shift={(\hdg:0.95)}]P) arc[start angle=\hdg, end angle=\hdg + \slip, radius=0.95]
    node[pos=0.6, right] {\(\beta\)};
  \draw[->, thick, rwthblue] (P) -- ++(\hdg + \slip:1.9) node[above] {\(v\)};
  \fill[white] (P) circle (1.8pt);
  \draw (P) circle (1.8pt);
  \node[above left, inner sep=2pt] at (P) {\(\bm p\)};

  % Global frame
  \draw[->] (-0.9, -1.5) -- ++(0.9, 0) node[right] {\(p_x\)};
  \draw[->] (-0.9, -1.5) -- ++(0, 0.9) node[above] {\(p_y\)};
\end{tikzpicture}
\caption{Kinematic bicycle model. \(\bm p = (p_x, p_y)\): position; \(\psi\): heading; \(v\): speed; \(\beta\): slip angle; \(\delta\): steering angle; \(\ell_{\mathrm r}\): rear-axle distance; \(\ell_{\mathrm{wb}}\): wheelbase.}
\label{fig:bicycle}
\end{figure}

Each CAV is modeled by the kinematic bicycle model \cite{rajamani2006vehicle} (\cref{fig:bicycle}). Vehicle \(i \in \mathcal I\) has the state \(\bm x^i = (\bm p^i, \psi^i, v^i, \delta^i) \in \mathcal X := \mathbb R^5\), with position \(\bm p^i = (p^i_x, p^i_y)\), heading \(\psi^i\), speed \(v^i \in [0, v_{\max}]\), and steering angle \(\delta^i \in [\delta_{\min}, \delta_{\max}]\). Its control \(\bm u^i = (a^i, \omega^i) \in \mathbb R^2\) contains acceleration \(a^i \in [a^i_{\min}, a^i_{\max}]\), \(a^i_{\min} < 0 < a^i_{\max} \le |a^i_{\min}|\), and steering rate \(\omega^i \in [\omega^i_{\min}, \omega^i_{\max}]\). The rear-axle distance \(\ell_{\mathrm r}\) and the wheelbase \(\ell_{\mathrm{wb}}\) define the slip angle \(\beta(\delta) = \arctan(\tfrac{\ell_{\mathrm r}}{\ell_{\mathrm{wb}}} \tan\delta)\). The dynamics of vehicle \(i\) are \(\dot{\bm x}^i = \bm f(\bm x^i, \bm u^i)\) with \(\bm f: \mathcal X \times \mathbb R^2 \to \mathbb R^5\) given by (we sometimes drop the superscript \(i\) where only one vehicle is involved)
\begin{equation*}
\begin{gathered}
\dot p_x = v\cos(\psi+\beta),\quad \dot p_y = v\sin(\psi+\beta),\\
\dot\psi = \tfrac{v}{\ell_{\mathrm r}}\sin\beta,\quad \dot v = a,\quad \dot\delta = \omega .
\end{gathered}
\end{equation*}
The controller runs at the sampling period \(\Delta t > 0\), and the vehicles evolve according to the explicit-Euler discretization
\begin{equation}\label{eq:euler}
\bm x^i_{k+1} = \bm x^i_k + \Delta t\, \bm f(\bm x^i_k, \bm u^i_k).
\end{equation}
The admissible controls \(\mathcal U^i(\bm x^i) \subseteq [a^i_{\min}, a^i_{\max}] \times [\omega^i_{\min}, \omega^i_{\max}]\) are those that keep the next speed and steering angle within their limits. We denote the joint state \(\bm x = (\bm x^1, \dots, \bm x^n) \in \mathcal X^n\), the joint control \(\bm u = (\bm u^1, \dots, \bm u^n) \in \mathbb R^{2n}\), the joint admissible set \(\mathcal U(\bm x) = \mathcal U^1(\bm x^1) \times \dots \times \mathcal U^n(\bm x^n)\), and the joint dynamics \(\bm x_{k+1} = \bm F(\bm x_k, \bm u_k)\), \(\bm F: \mathcal X^n \times \mathbb R^{2n} \to \mathcal X^n\).

Safety depends on the area that each vehicle occupies, which is determined by its pose. The pose of vehicle \(i\) is \(\bm\xi^i = \bm\xi(\bm x^i) = (\bm p^i, \psi^i)\), with \(\bm\xi: \mathcal X \to \mathcal P := \mathbb R^2 \times \mathbb S^1\), where \(\mathbb S^1\) is the unit circle of headings. At pose \(\bm\xi^i\), vehicle \(i\) occupies the rectangle \(\mathcal R(\bm\xi^i) \subset \mathbb R^2\) with its half-length \(\ell^i_{\mathrm{lon}}\) and half-width \(\ell^i_{\mathrm{lat}}\). We denote \(\bm\xi^i_k = \bm\xi(\bm x^i_k)\) for the pose at step \(k\) and \(\bm\xi(\bm x) = (\bm\xi(\bm x^1), \dots, \bm\xi(\bm x^n)) \in \mathcal P^n\) for the joint pose.

The nominal controls come from a MARL policy. At step \(k\), vehicle \(i\) receives a local observation \(\bm o^i_k\) from the observation space \(\mathcal O\), and \(\bm o_k = (\bm o^1_k, \dots, \bm o^n_k)\) is the joint observation. The policy \(\pi_{\bm\theta}: \mathcal O \to [0, v_{\max}] \times [\delta_{\min}, \delta_{\max}]\) with parameters \(\bm\theta\) maps \(\bm o^i_k\) to a nominal action, which consists of a reference speed \(v^i_{\mathrm{ref},k}\) and a reference steering angle \(\delta^i_{\mathrm{ref},k}\). This action is converted to the nominal control \(\bm u^i_{\mathrm{nom},k} = \big((v^i_{\mathrm{ref},k} - v^i_k)/\Delta t,\ (\delta^i_{\mathrm{ref},k} - \delta^i_k)/\Delta t\big)\), saturated to \(\mathcal U^i(\bm x^i_k)\), so that the next speed and steering angle approach the references as closely as one step allows.

For safety, each vehicle must keep a Euclidean clearance of at least \(\varepsilon_{\mathrm{lane}} > 0\) from the left and right boundaries of its route and of at least \(\varepsilon_{\mathrm{veh}} > 0\) from every other vehicle. We call each of these requirements a \emph{safety constraint} \(e \in \mathcal E\), where \(\mathcal E\) is the set of all constraints. A constraint is either a lane constraint for one boundary side \(s \in \{\mathrm L, \mathrm R\}\) (left, right) of one vehicle \(i\), or a vehicle constraint for a pair \((i, j)\) of vehicles. We denote by \(\mathcal E^i_{\mathrm L}, \mathcal E^i_{\mathrm R}, \mathcal E^i_{\mathrm V} \subseteq \mathcal E\) the left-side, right-side, and vehicle constraints that involve vehicle \(i\). Each constraint \(e\) is encoded by a safety margin \(h_e: \mathcal P^n \to \mathbb R\), and a nonnegative margin implies the required clearance. The safe set is thus the set of joint states at which all safety margins are nonnegative,
\begin{equation*}
\mathcal S = \{\bm x \in \mathcal X^n : h_e(\bm\xi(\bm x)) \ge 0,\ \forall e \in \mathcal E\}.
\end{equation*}

The goal is to design a safety filter that maps the nominal controls \(\bm u_{\mathrm{nom},k}\) of the MARL policy to admissible controls \(\bm u_k \in \mathcal U(\bm x_k)\) such that the joint state stays in the safe set, \(\bm x_k \in \mathcal S\) for all \(k \ge 0\) under the assumptions below. The filtered controls should deviate minimally from the nominal ones and never lead to a state from which a collision is unavoidable, so that the filter always finds a safe control (recursive feasibility). Safety and recursive feasibility should also hold under fully decentralized control, where each vehicle computes its own control from local information only. In addition, the filter should help train and fine-tune the MARL policy, so that the policy itself becomes safer and the filter intervenes less.

We make the following assumptions.
\begin{enumerate}[label=(A\arabic*), ref=(A\arabic*), leftmargin=*]
\item\label{as:plant}
  The vehicles evolve according to \eqref{eq:euler}, and all guarantees refer to this model at the sampling instants.
\item\label{as:init}
  The joint initial state is in the safe set \(\mathcal S\), and collisions are still avoidable from it. Any vehicle that appears later preserves this property of the joint state.
\item\label{as:coop}
  All CAVs are cooperative and can share their states, routes, and vehicle parameters via reliable, delay-free vehicle-to-everything communication.
\end{enumerate}

\section{Backup-Horizon Control Barrier Functions}\label{sec:bh-cbf}

This section presents our BH-CBF and the fully decentralized safety filter built on it. \Cref{sec:margins} defines the safety margins on the exact geometry of rectangular vehicles and polyline lane boundaries. \Cref{sec:one-step} shows that the standard one-step CBF condition on these margins can become infeasible under control bounds. To restore recursive feasibility, \cref{sec:backup} introduces our backup policy, which brakes every vehicle to a standstill along its route. We predict the poses along this policy until all vehicles have stopped. \Cref{sec:centralized} defines the backup value as the smallest margin along this prediction and imposes the CBF condition on it. This condition is stated for the joint system of all vehicles. \Cref{sec:decentralized} derives our decentralized filter by splitting the pair condition between the two vehicles, and \cref{sec:filter} solves the resulting nonconvex filter problems by QPs whose output is certified by an exact check.

\subsection{Margins With Exact Vehicle and Road Geometry}\label{sec:margins}

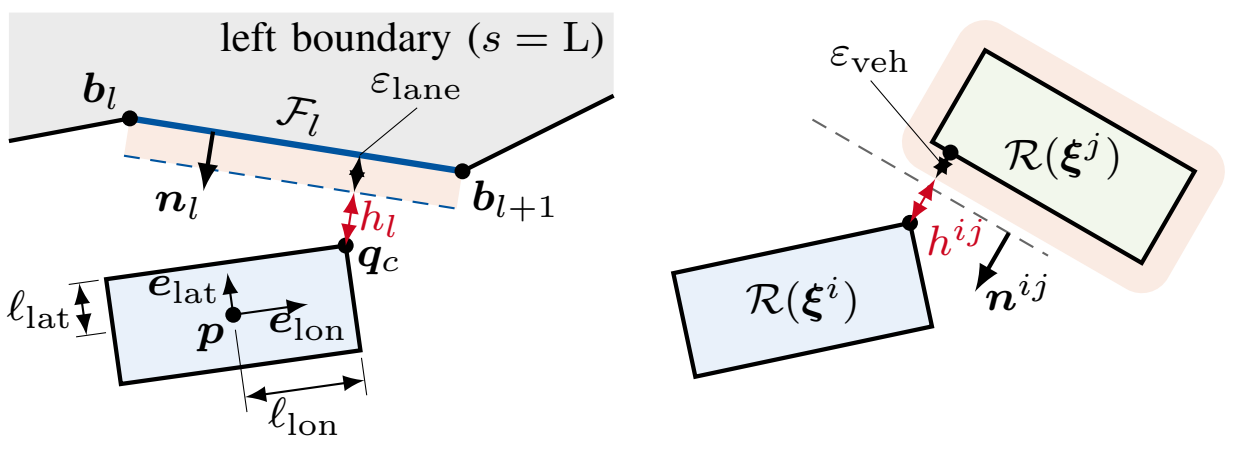
\begin{figure}[t]
\centering
% Margin geometry: (a) lane margin, (b) pair margin; notation as in the text.
% (a) facet F_l from b_l = (0.4, 1.35) to b_{l+1} = (2.6, 1.0), inward normal n_l = (-0.1571, -0.9876);
%     vehicle at p = (1.084, 0.051), heading 8 deg, half-length 0.8, half-width 0.35; clearance 0.25.
% (b) rectangles at headings 12 and -30 deg (vehicle j green); closest points (0.710, 0.509) on A (corner) and
%     (0.978, 0.974) on B (edge, normal (-0.5, -0.866)); clearance 0.2, exaggerated like in (a).
\begin{tikzpicture}[
    font=\footnotesize,
    >=latex,
    veh/.style={draw, thick, fill=rwthblue10},
    pt/.style={circle, fill, inner sep=0pt, minimum size=3pt},
    dim/.style={<->, thin},
    ext/.style={very thin},
    lbl/.style={inner sep=1.5pt},
  ]
  % ---------------- (a) Lane margin ----------------
  % Off-road area and left boundary polyline
  \fill[black!8] (-0.4, 1.2) -- (0.4, 1.35) -- (2.6, 1.0) -- (3.6, 1.5) -- (3.6, 2.05) -- (-0.4, 2.05) -- cycle;
  % Clearance zone (forbidden, red as in the other figures)
  \fill[rwthred10] (0.4, 1.35) -- (2.6, 1.0) -- (2.561, 0.753) -- (0.361, 1.103) -- cycle;
  \draw[thick] (-0.4, 1.2) -- (0.4, 1.35) -- (2.6, 1.0) -- (3.6, 1.5);
  \draw[very thick, rwthblue] (0.4, 1.35) -- node[lbl, above, sloped, black] {\(\mathcal F_l\)} (2.6, 1.0);
  \node[pt] at (0.4, 1.35) {};
  \node[pt] at (2.6, 1.0) {};
  \node[lbl, above left] at (0.4, 1.35) {\(\bm b_l\)};
  \node[lbl, below right] at (2.6, 1.0) {\(\bm b_{l+1}\)};
  \node[lbl, anchor=north east] at (3.6, 2.05) {left boundary (\(s = \mathrm L\))};
  % Clearance line at distance eps_lane from the facet (0.25, exaggerated so that its dimension arrow is visible)
  \draw[densely dashed, rwthblue] (0.361, 1.103) -- (2.561, 0.753);
  % Inward unit normal
  \draw[->, thick] (0.95, 1.2625) -- (0.887, 0.868) node[lbl, below left, pos=1, xshift=0.6mm, yshift=0.6mm] {\(\bm n_l\)};

  % Vehicle rectangle in its body frame
  \begin{scope}[shift={(1.084, 0.051)}, rotate=8]
    \draw[veh] (-0.8, -0.35) rectangle (0.8, 0.35);
    \draw[->] (0, 0) -- (0.5, 0) node[lbl, below] {\(\bm e_{\mathrm{lon}}\)};
    \draw[->] (0, 0) -- (0, 0.28) node[lbl, left, pos=0.6] {\(\bm e_{\mathrm{lat}}\)};
    \node[pt] at (0, 0) {};
    \node[lbl, below left] at (0, 0) {\(\bm p\)};
    \node[pt] at (0.8, 0.35) {};
    \node[lbl, anchor=west] at (0.8, 0.25) {\(\bm q_c\)};
    % Half-length and half-width
    \draw[ext] (0, -0.1) -- (0, -0.65);
    \draw[ext] (0.8, -0.4) -- (0.8, -0.65);
    \draw[dim] (0, -0.55) -- node[lbl, below] {\(\ell_{\mathrm{lon}}\)} (0.8, -0.55);
    \draw[ext] (-0.85, 0) -- (-1.05, 0);
    \draw[ext] (-0.85, 0.35) -- (-1.05, 0.35);
    \draw[dim] (-0.98, 0) -- node[lbl, left] {\(\ell_{\mathrm{lat}}\)} (-0.98, 0.35);
  \end{scope}

  % Corner-to-facet distance: margin h_l plus eps_lane
  \draw[dim, rwthred] (1.828, 0.509) -- node[lbl, right] {\(h_l\)} (1.883, 0.861);
  \draw[dim] (1.883, 0.861) -- (1.922, 1.108);
  \draw[ext] (1.922, 1.108) -- (2.30, 1.42) node[lbl, above] {\(\varepsilon_{\mathrm{lane}}\)};

  \node at (1.6, -1.05) {(a)};

  % ---------------- (b) Pair margin ----------------
  \begin{scope}[xshift=4.85cm, yshift=0.15cm]
    \begin{scope}[rotate=12]
      \draw[veh] (-0.8, -0.35) rectangle (0.8, 0.35);
      \node {\(\mathcal R(\bm\xi^i)\)};
    \end{scope}
    \begin{scope}[shift={(1.725, 0.947)}, rotate=-30]
      % Clearance zone around R(xi^j): points closer than eps_veh (Euclidean, so rounded corners)
      \fill[rwthred10, rounded corners=0.2cm] (-1.0, -0.55) rectangle (1.0, 0.55);
      \draw[veh, fill=rwthgreen10] (-0.8, -0.35) rectangle (0.8, 0.35);
      \node {\(\mathcal R(\bm\xi^j)\)};
    \end{scope}
    % Separating line orthogonal to the separating normal, through the midpoint of the closest points
    \draw[densely dashed, black!60] (0.065, 1.192) -- (1.623, 0.292);
    % Signed distance between the closest points: margin h^{ij} plus eps_veh
    \node[pt] at (0.710, 0.509) {};
    \node[pt] at (0.978, 0.974) {};
    \draw[dim, rwthred] (0.710, 0.509) -- node[lbl, below right, pos=0.3] {\(h^{ij}\)} (0.878, 0.801);
    \draw[dim] (0.878, 0.801) -- (0.978, 0.974);
    \draw[ext] (0.928, 0.888) -- (0.45, 1.45) node[lbl, above] {\(\varepsilon_{\mathrm{veh}}\)};
    % Separating normal n^{ij}, pointing from R(xi^j) to R(xi^i)
    \draw[->, thick] (1.364, 0.442) -- (1.139, 0.052) node[lbl, right] {\(\bm n^{ij}\)};
    \node at (0.85, -1.2) {(b)};
  \end{scope}
\end{tikzpicture}
\caption{Margin geometry. (a) Facet margin \(h_l\), whose minimum over all facets is the lane margin. \(\bm p\): position; \(\bm e_{\mathrm{lon}}, \bm e_{\mathrm{lat}}\): body axes; \(\bm q_c\): corner; \(\ell_{\mathrm{lon}}, \ell_{\mathrm{lat}}\): half-length, half-width; \(\mathcal F_l = [\bm b_l, \bm b_{l+1}]\): facet of the left boundary (\(s = \mathrm L\)); \(\bm n_l\): inward normal; red: clearance \(\varepsilon_{\mathrm{lane}}\). (b) Pair margin \(h^{ij}\). Red: clearance \(\varepsilon_{\mathrm{veh}}\); \(\bm n^{ij}\): separating normal; dashed: separating line.}
\label{fig:margins}
\end{figure}

This section defines the lane margin of a vehicle to each boundary of its route and the pair margin of two vehicles. Both are built from the corners of the vehicle rectangles.

For a pose \(\bm\xi = (\bm p, \psi) \in \mathcal P\), let \(\bm e_{\mathrm{lon}} = (\cos\psi, \sin\psi)\) and \(\bm e_{\mathrm{lat}} = (-\sin\psi, \cos\psi)\) be the body axes and \(\bm q_c: \mathcal P \to \mathbb R^2\), \(c = 1, \dots, 4\), the corners of \(\mathcal R(\bm\xi)\), one for each of the four sign combinations in \(\bm p \pm \ell_{\mathrm{lon}} \bm e_{\mathrm{lon}} \pm \ell_{\mathrm{lat}} \bm e_{\mathrm{lat}}\) (see \(\bm e_{\mathrm{lon}}\), \(\bm e_{\mathrm{lat}}\), and \(\bm q_c\) in \cref{fig:margins}a). The \emph{lane margin} measures how far a vehicle stays from the boundaries of its route. The route of vehicle \(i\) has a left and a right boundary, indexed by the boundary side \(s \in \{\mathrm L, \mathrm R\}\). The boundary of side \(s\) is a polyline whose line segments, also called facets, are indexed by \(l \in \mathcal L^i_s\) (left boundary in \cref{fig:margins}a, with facet \(\mathcal F_l\) in blue). Each facet \(\mathcal F_l = [\bm b_l, \bm b_{l+1}]\) is the segment between the consecutive vertices \(\bm b_l, \bm b_{l+1} \in \mathbb R^2\) and has the unit normal \(\bm n_l\) pointing into the lane. Let \(\sigma_{\mathrm L} = +1\) and \(\sigma_{\mathrm R} = -1\), so that \(\sigma_s \bm e_{\mathrm{lat}}\) points from the vehicle toward the boundary of side \(s\). For each facet, we check four separating directions: the facet normal \(\bm n_l\) and the three body directions \(\bm e_{\mathrm{lon}}\), \(-\bm e_{\mathrm{lon}}\), and \(\sigma_s \bm e_{\mathrm{lat}}\). Along each direction, we take the gap between the projections of the rectangle and the facet, minus \(\varepsilon_{\mathrm{lane}}\), with the facet endpoints \(\bm b \in \{\bm b_l, \bm b_{l+1}\}\):
\begin{enumerate*}[label=(\roman*)]
\item
  facet normal \(\bm n_l\): \(\ \min_c \langle \bm n_l, \bm q_c(\bm\xi) - \bm b_l \rangle - \varepsilon_{\mathrm{lane}}\);
\item
  front \(\bm e_{\mathrm{lon}}\): \(\ \min_{\bm b} \langle \bm e_{\mathrm{lon}}, \bm b - \bm p \rangle - \ell_{\mathrm{lon}} - \varepsilon_{\mathrm{lane}}\);
\item
  rear \(-\bm e_{\mathrm{lon}}\): \(\ -\max_{\bm b} \langle \bm e_{\mathrm{lon}}, \bm b - \bm p \rangle - \ell_{\mathrm{lon}} - \varepsilon_{\mathrm{lane}}\); and
\item
  boundary side \(\sigma_s \bm e_{\mathrm{lat}}\): \(\ \min_{\bm b} \sigma_s \langle \bm e_{\mathrm{lat}}, \bm b - \bm p \rangle - \ell_{\mathrm{lat}} - \varepsilon_{\mathrm{lane}}\).
\end{enumerate*}
The largest of (i)--(iv) is a separating-axis bound, which we take as the facet margin \(h_l: \mathcal P \to \mathbb R\). The lane margin \(h^i_s: \mathcal P \to \mathbb R\) of side \(s\) is the smallest facet margin,
\begin{equation*}
h^i_s(\bm\xi) = \min_{l \in \mathcal L^i_s} h_l(\bm\xi).
\end{equation*}
A projection gap never exceeds the distance between two sets. Hence, the facet margin \(h_l(\bm\xi)\) is a lower bound on the Euclidean distance between the rectangle \(\mathcal R(\bm\xi)\) and the facet \(\mathcal F_l\), minus \(\varepsilon_{\mathrm{lane}}\). Thus, \(h^i_s(\bm\xi) \ge 0\) implies a clearance of at least \(\varepsilon_{\mathrm{lane}}\) to every facet of side \(s\).

The \emph{pair margin} builds on the signed distance between two rectangles. Consider two rectangles \(\mathcal A = \mathcal R(\bm\xi)\) and \(\mathcal B = \mathcal R(\bm\xi')\) with corners \(\bm q_c(\mathcal A) = \bm q_c(\bm\xi)\) and \(\bm q_c(\mathcal B) = \bm q_c(\bm\xi')\), \(c = 1, \dots, 4\). As for the lane margin, their gap along a unit vector \(\bm w \in \mathbb R^2\) is the gap between their projections onto \(\bm w\), \(\mathrm{gap}_{\bm w}(\mathcal A, \mathcal B) = \min_c \langle \bm w, \bm q_c(\mathcal A)\rangle - \max_c \langle \bm w, \bm q_c(\mathcal B)\rangle \in \mathbb R\). Their signed distance is the largest gap over all directions,
\begin{equation}
\mathrm{sd}(\mathcal A, \mathcal B) = \max_{\|\bm w\| = 1} \mathrm{gap}_{\bm w}(\mathcal A, \mathcal B) \in \mathbb R.\label{eq:sd}
\end{equation}
We call the unit vector \(\bm w\) that attains the maximum in \eqref{eq:sd} the \emph{separating normal}. It points from \(\mathcal B\) to \(\mathcal A\) (\(\bm n^{ij}\) in \cref{fig:margins}b, orthogonal to the dashed separating line). For disjoint rectangles, \(\mathrm{sd}\) is their Euclidean distance and the separating normal is unique; for overlapping ones, \(\mathrm{sd}\) is minus their penetration depth. Both \(\mathrm{sd}\) and the separating normal have a closed form. With the signed distance, the margin of a pair of vehicles is \(h^{ij}: \mathcal P^2 \to \mathbb R\), \(h^{ij}(\bm\xi^i, \bm\xi^j) = \mathrm{sd}(\mathcal R(\bm\xi^i), \mathcal R(\bm\xi^j)) - \varepsilon_{\mathrm{veh}}\), so \(h^{ij} \ge 0\) holds if and only if the rectangles keep a Euclidean clearance of at least \(\varepsilon_{\mathrm{veh}}\). In contrast, the CBF for polygons in \cite{wu2025optimizationfree} smooths a lower bound on the signed distance. For example, for two rectangles, this bound restricts the maximum in \(\mathrm{sd}\) to the eight signed body axes and underestimates a positive clearance. For two aligned rectangles offset diagonally by \SI{1}{\meter} along both axes, it gives \SI{1}{\meter} instead of \SI{1.41}{\meter}.

The margin \(h_e\) of constraint \(e \in \mathcal E\) evaluates \(h^i_s\) at \(\bm\xi^i\) for boundary side \(s\) of vehicle \(i\), and \(h^{ij}\) at \((\bm\xi^i, \bm\xi^j)\) for a pair. These margins are not differentiable everywhere. Our discrete-time conditions do not need derivatives because they compare margin values at predicted poses.

\subsection{Exact Euler Prediction and the One-Step Condition}\label{sec:one-step}

The Euler step \eqref{eq:euler} updates the pose with the current speed and steering angle. Hence the predicted pose \(\bm\xi_{k+1|k} = \bm\xi(\bm F(\bm x_k, \bm u_k)) \in \mathcal P^n\) at \(k+1\) does not depend on the current control \(\bm u_k\). The predicted pose \(\bm\xi_{k+2|k}(\bm u) \in \mathcal P^n\) at \(k+2\) for a candidate control \(\bm u\) depends on \(\bm u\) only through the speed \(v^i_{k+1}\) and steering angle \(\delta^i_{k+1}\) of each vehicle, and not on the control at \(k+1\). Both are closed-form functions of \(\bm x_k\). The current control thus first affects the pose two steps ahead. Applying the standard discrete-time CBF condition \cite{agrawal2017discrete} to the margin \(h_e(\bm\xi_{k+1|k})\) of the next pose as a function of \(\bm x_k\) gives the \emph{one-step condition}
\begin{equation}
h_e\big(\bm\xi_{k+2|k}(\bm u)\big) \ge (1-\lambda)\, h_e\big(\bm\xi_{k+1|k}\big),\label{eq:one-step}
\end{equation}
with decay rate \(\lambda \in (0, 1]\).
The TTCBF \cite{xu2025highorder,xu2026ttcbf} imposes this condition on a second-order Taylor prediction over \(2\Delta t\), in which the acceleration \(a_k\) moves the pose by \(2\Delta t^2 a_k\). Our exact Euler prediction moves it by \(\Delta t^2 a_k\), since \(a_k\) changes the speed \(v_{k+1}\) by \(\Delta t\, a_k\), which acts on the pose for one step.

If condition \eqref{eq:one-step} holds at every step, it keeps the margins nonnegative by induction \cite{agrawal2017discrete}, since \(\bm\xi_{k+2|k}(\bm u_k) = \bm\xi_{k+2|k+1}\). However, satisfying it at step \(k\) does not ensure that it can be satisfied at step \(k+1\). The condition looks only two steps ahead and allows the margin to shrink by the fraction \(\lambda\) per step, regardless of the speed. A vehicle with bounded deceleration, in contrast, needs a stopping distance that grows quadratically with its speed. Consider a vehicle that approaches a stopped vehicle. While the margin is large, \eqref{eq:one-step} lets the vehicle keep its speed, because the allowed shrinkage exceeds the distance it covers in one step. The condition starts to demand braking only when the margin has become small, and then the stopping distance may already exceed the margin. In this case, \eqref{eq:one-step} still holds at step \(k\) with mild braking, but even maximum braking violates it at step \(k+1\), and the QP becomes infeasible. Hence \(\mathcal S\) is only a candidate safe set for \eqref{eq:one-step} because it contains states from which no admissible control keeps the margins nonnegative. We restore recursive feasibility by looking ahead until the vehicles can stop safely, using the backup policy defined in the next section.

\subsection{Backup Policy}\label{sec:backup}

\begin{figure}[t]
\centering
% Backup policy, backup prediction, and stopping horizon; notation as in the text.
% Schematic Euler rollout of one vehicle (figure units: cm) in the route frame (arc length, lateral offset,
% relative heading), unicycle with curvature c in place of the steering angle, mapped onto a left arc of
% radius 6 around (0, 5.9) (the route centerline; boundaries at +-0.7). The vehicle starts 0.1 left of the
% centerline at (0, 0), parallel to it, with the curvature of the road. Displacement 0.18 cm per speed unit and
% step; v_k = 5, a_max dt = 0.5, |a_min| dt = 1, so N = ceil(5 / 1) + 2 = 7. Backup: a = max(-1, -v); its
% curvatures (a schematic stand-in for pure pursuit) are the smoothest sequence (least squared change per
% step) that reaches the centerline, aligned with it, at standstill; for gray, the margin also stays above its
% final value. Gray curvatures (t = 1..4): -0.081 0.505 0.167 0.167. Blue: xi_{k+t|k}(u) for
% u = (a_max, curvature change -0.20), backup curvatures (t = 2..6): 0.321 0.109 -0.030 -0.069 -0.069.
% (c) Margin to the outer (right) boundary, h = 6.7 - max corner distance to the arc center:
%     gray (t = 1..5) 0.582 0.506 0.484 0.484 0.484, so B_e(x_k) = 0.484;
%     blue (t = 1..7) 0.582 0.498 0.418 0.376 0.430 0.473 0.484; lambda = 0.5. The axis starts at h = 0.
% Labels use \contour (package contour, loaded in 00_ps.sty) for a white outline instead of a white box.
\begin{tikzpicture}[
    font=\footnotesize,
    >=latex,
    veh/.style={draw, thick, fill=rwthblue10},
    pt/.style={circle, fill, inner sep=0pt, minimum size=2.5pt},
    lbl/.style={inner sep=1.5pt},
  ]
  % Label text with a white outline that keeps lines readable behind it
  \newcommand{\halo}[1]{\contour{white}{#1}}
  % Rectangle of half-length 0.45 and half-width 0.2 at x, y, heading; #4: style
  \newcommand{\car}[4]{\draw[#4] (#1, #2) ++(#3:0.45) ++(#3+90:0.2) -- ++(#3+180:0.9)
    -- ++(#3-90:0.4) -- ++(#3:0.9) -- cycle;}

  % ---------------- (a) Tail prediction ----------------
  % Arcs outside the bounding box (their control points would add blank space above the figure)
  \path (-0.55, -0.9) (4.45, 2.05);
  \begin{pgfinterruptboundingbox}
  \draw[densely dashed, black!40] (0, 5.9) ++(-94.5:6) arc[start angle=-94.5, end angle=-46, radius=6];
  \draw[thin] (0, 5.9) ++(-94.5:6.7) arc[start angle=-94.5, end angle=-49, radius=6.7];
  \draw[thin] (0, 5.9) ++(-94.5:5.3) arc[start angle=-94.5, end angle=-48, radius=5.3];
  \end{pgfinterruptboundingbox}

  % Backup prediction (gray)
  \draw[black!45] (0.897, 0.069) -- (1.600, 0.219) -- (2.139, 0.294) -- (2.471, 0.433) -- (2.634, 0.509);
  \foreach \x/\y in {1.600/0.219, 2.139/0.294, 2.471/0.433, 2.634/0.509}
    \node[pt, black!45] at (\x, \y) {};
  \car{2.634}{0.509}{26.0}{black!45, thick}
  \node[lbl, text=black!55, anchor=west] at (3.16, 0.27) {\halo{\(\bar{\bm\xi}^i_{k+N|k}\)}};
  \node[lbl, text=black!55, align=center, anchor=north, inner sep=0.5pt] (brktxt) at (2.45, 0.0)
    {\halo{backup}\\[-1pt]\halo{prediction}};
  \draw[very thin, black!55] (brktxt.north) -- (2.468, 0.40);  % to the dot of t = 4

  % Prediction for the control u (blue)
  \draw[rwthblue] (0.897, 0.069) -- (1.859, 0.298) -- (2.664, 0.450) -- (3.215, 0.756) -- (3.560, 1.043) -- (3.761, 1.222) -- (3.829, 1.281);
  \foreach \x/\y in {1.859/0.298, 2.664/0.450, 3.215/0.756, 3.560/1.043, 3.761/1.222, 3.829/1.281}
    \node[pt, rwthblue] at (\x, \y) {};
  \car{1.859}{0.298}{6.7}{rwthblue}
  \car{3.829}{1.281}{39.7}{rwthblue, thick}
  \node[lbl, text=rwthblue, anchor=south] at (1.80, 0.55) {\halo{\(\bm\xi^i_{k+2|k}\)}};
  \node[lbl, text=rwthblue, align=center, anchor=east, inner sep=0.5pt] (nomtxt) at (2.55, 1.45)
    {\halo{candidate}\\[-1pt]\halo{prediction}};
  \draw[very thin, rwthblue] (nomtxt.east) -- (3.20, 0.78);
  \node[lbl, text=rwthblue, anchor=south] at (3.55, 1.6) {\halo{\(\bm\xi^i_{k+N|k}\)}};

  % Current pose and the predicted pose at k+1, which the control cannot change
  \car{0}{0}{0}{veh}
  \car{0.897}{0.069}{8.6}{thick}
  \node[pt] at (0, 0) {};
  \node[pt] at (0.897, 0.069) {};
  \node[lbl, anchor=north] at (0, -0.22) {\halo{\(\bm\xi^i_k\)}};
  \node[lbl, anchor=north] at (0.95, -0.15) {\halo{\(\bm\xi^i_{k+1|k}\)}};

  \node at (2.0, -1.25) {(a)};

  % ---------------- (b) Speed along the prediction ----------------
  \begin{scope}[xshift=5.6cm, yshift=-0.75cm, x=0.27cm, y=0.3cm]
    \draw[->] (0, 0) -- (8.3, 0) node[lbl, below] {\(t\)};
    \draw[->] (0, 0) -- (0, 6.4) node[lbl, right] {speed};
    \foreach \t in {0, 1, 2} \draw (\t, 0) -- (\t, -0.2) node[lbl, below] {\(\t\)};
    \draw (7, 0) -- (7, -0.2) node[lbl, below] {\(N\)};
    \draw[densely dashed, very thin] (7, 0) -- (7, 5.9);
    \draw (0, 5) -- (-0.2, 5) node[lbl, left] {\(v^i_k\)};
    \draw[densely dotted, very thin] (0, 5) -- (8.1, 5);
    \draw[black!45] (0, 5) -- (1, 4) -- (2, 3) -- (3, 2) -- (4, 1) -- (5, 0) -- (8.1, 0);
    \foreach \t/\v in {1/4, 2/3, 3/2, 4/1, 5/0} \node[pt, black!45] at (\t, \v) {};
    \draw[rwthblue] (0, 5) -- (1, 5.5) -- (2, 4.5) -- (3, 3.5) -- (4, 2.5) -- (5, 1.5) -- (6, 0.5) -- (7, 0)
      -- (8.1, 0);
    \foreach \t/\v in {1/5.5, 2/4.5, 3/3.5, 4/2.5, 5/1.5, 6/0.5, 7/0}
      \node[pt, rwthblue] at (\t, \v) {};
    \node[pt] at (0, 5) {};
  \end{scope}
  \node at (6.7, -1.25) {(b)};

  % ---------------- (c) Margin along the predictions ----------------
  \begin{scope}[xshift=0.2cm, yshift=-3.95cm, x=0.7cm, y=2.9cm]
    \fill[rwthred10] (0, 0) rectangle (7.9, 0.242);
    \draw[->] (0, 0) -- (8.2, 0) node[lbl, below] {\(t\)};
    \draw[->] (0, 0) -- (0, 0.68) node[lbl, right] {\(h_e\)};
    \foreach \t in {0, 1, 2} \draw (\t, 0) -- (\t, -0.021) node[lbl, below] {\(\t\)};
    \draw (7, 0) -- (7, -0.021) node[lbl, below] {\(N\)};
    % Backup values and the bound of the centralized condition
    \draw[rwthblue, densely dashed, thin] (4, 0.376) -- (7.9, 0.376) node[lbl, right, text=black] {\(B_e(\bm x_{k+1})\)};
    \draw[rwthred] (0, 0.242) -- (7.9, 0.242) node[lbl, right] {\((1-\lambda) B_e(\bm x_k)\)};
    % Backup prediction made at k (gray)
    \draw[black!45] (1, 0.582) -- (2, 0.506) -- (3, 0.484) -- (4, 0.484) -- (5, 0.484) -- (7.9, 0.484) node[lbl, right, text=black] {\(B_e(\bm x_k)\)};
    \foreach \t/\h in {1/0.582, 2/0.506, 3/0.484, 4/0.484, 5/0.484} \node[pt, black!45] at (\t, \h) {};
    % Prediction for u (blue)
    \draw[rwthblue] (1, 0.582) -- (2, 0.498) -- (3, 0.418) -- (4, 0.376) -- (5, 0.430) -- (6, 0.473) -- (7, 0.484);
    \foreach \t/\h in {1/0.582, 2/0.498, 3/0.418, 4/0.376, 5/0.430, 6/0.473, 7/0.484}
      \node[pt, rwthblue] at (\t, \h) {};
    % Minimum of the blue curve over t >= 2, which is B_e(x_{k+1})
    \draw[rwthblue, thick] (4, 0.376) circle[radius=3.2pt];
  \end{scope}
  \node at (3.6, -4.6) {(c)};
\end{tikzpicture}
\caption{Prediction of vehicle \(i\) at step \(k\) over \(t\) (schematic). Blue: candidate prediction \(\bm\xi^i_{k+t|k}(\bm u^i)\) for a control \(\bm u^i\) that accelerates and steers away from the route; gray: backup prediction \(\bar{\bm\xi}^i_{k+t|k}\), which tracks the route centerline (dashed). (a) Poses; \(\bm\xi^i_{k+1|k}\) is the same for both predictions. (b) Speed. (c) Margin \(h_e\) of the right lane constraint; \eqref{eq:bh-cbf} keeps the blue curve above \((1-\lambda) B_e(\bm x_k)\) from \(t = 2\) on.}
\label{fig:backup}
\end{figure}
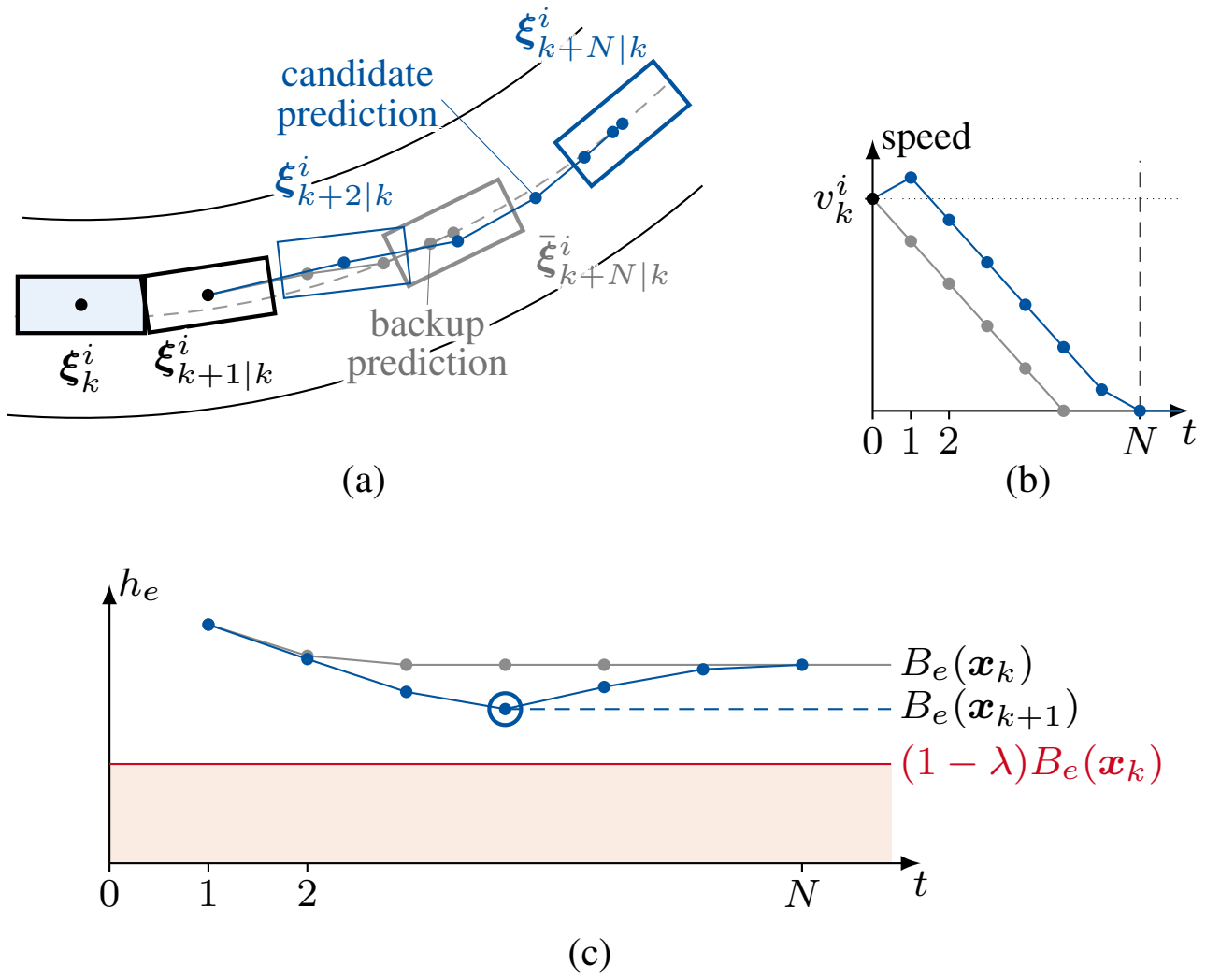

To look beyond the next step, we predict the vehicles along a backup policy that is always admissible.

\begin{definition}[Backup Policy]\label{def:backup}
The backup policy \(\bm\kappa^i: \mathcal X \to \mathbb R^2\) of vehicle \(i\) applies maximum braking until standstill while steering along its route centerline. It accelerates with \(a^i = \min(a^i_{\max}, \max(a^i_{\min}, -v^i/\Delta t))\) and steers by pure pursuit toward a look-ahead point on the centerline, within the steering-rate and steering-angle limits. The joint backup policy is \(\bm\kappa: \mathcal X^n \to \mathbb R^{2n}\), \(\bm\kappa(\bm x) = (\bm\kappa^1(\bm x^1), \dots, \bm\kappa^n(\bm x^n))\).
\end{definition}
For short, we also call \(\bm\kappa\) the backup. Our backup steers toward the centerline so that a braking vehicle stays clear of the lane boundaries and of vehicles in other lanes. It has the following three properties:

\begin{enumerate}[label=(K\arabic*), ref=(K\arabic*), leftmargin=*]
\item\label{k:admissible}
  The backup is admissible: \(\bm\kappa(\bm x) \in \mathcal U(\bm x)\).
\item\label{k:decoupled}
  The backup is decoupled and memoryless: the backup control \(\bm\kappa^i(\bm x^i)\) of vehicle \(i\) depends only on its own current state and route, not on the states of other vehicles or on past states.
\item\label{k:stopping}
  The backup stops: the speed reaches zero after finitely many steps and stays zero.
\end{enumerate}
The backup rollout \(\bm\varphi_t: \mathcal X^n \to \mathcal X^n\) maps a state \(\bm x\) to the state after \(t \ge 0\) steps of the backup, with \(\bm\varphi_0(\bm x) = \bm x\). By \ref{k:decoupled}, it decomposes into the single-vehicle rollouts \(\bm\varphi^i_t: \mathcal X \to \mathcal X\), and \(t\) backup steps from \(\bm\varphi_1(\bm x)\) lead to \(\bm\varphi_{t+1}(\bm x)\).

With the backup, we predict future poses at step \(k\) if a candidate control \(\bm u\) is applied now and every vehicle follows the backup afterwards:
\begin{equation*}
\bm\xi_{k+t|k}(\bm u) = \bm\xi\big(\bm\varphi_{t-1}(\bm F(\bm x_k, \bm u))\big), \qquad t = 1, 2, \dots,
\end{equation*}
with \(\bm\xi_{k+t|k}: \mathbb R^{2n} \to \mathcal P^n\) for the given \(\bm x_k\). We call \(\bm\xi_{k+t|k}(\bm u)\) the \emph{candidate prediction}. By \ref{k:decoupled}, the component \(\bm\xi^i_{k+t|k}(\bm u^i)\) of vehicle \(i\) depends only on \(\bm x^i_k\) and \(\bm u^i\). The control \(\bm u = \bm\kappa(\bm x_k)\) gives the \emph{backup prediction} \(\bar{\bm\xi}_{k+t|k} = \bm\xi(\bm\varphi_t(\bm x_k))\), in which every vehicle follows the backup from step \(k\) on (gray in \cref{fig:backup}). Since the backup stops every vehicle \ref{k:stopping}, a finite stopping horizon covers both predictions.

\begin{lemma}[Stopping Horizon]\label{lem:stopping}
Let \(N \in \mathbb N\) satisfy \(N \ge \max_{i \in \mathcal I} \lceil v^i_k / (|a^i_{\min}| \Delta t) \rceil + 2\). Then every vehicle stands still from step \(k+N\) on in the candidate prediction of every admissible \(\bm u\), so \(\bm\xi_{k+t|k}(\bm u) = \bm\xi_{k+N|k}(\bm u)\) for all \(t \ge N\). The same holds for the backup prediction.
\end{lemma}

\emph{Proof.} The first control raises the speed of vehicle \(i\) to at most \(v^i_k + a^i_{\max}\Delta t \le v^i_k + |a^i_{\min}|\Delta t\) because \(a^i_{\max} \le |a^i_{\min}|\) (\cref{sec:problem}). Hence the backup from \(\bm x^i_{k+1}\) reaches zero speed after at most \(\lceil v^i_k / (|a^i_{\min}|\Delta t) \rceil + 1\) steps, i.e., at step \(k+N\) at the latest; the last step only sets the speed to zero (last blue segment in \cref{fig:backup}b), after which the pose is constant by \ref{k:stopping} and \eqref{eq:euler}. The backup prediction brakes from \(v^i_k\) one step earlier and thus stops no later. \(\square\)

The stopping horizon replaces the fixed horizon and the terminal constraint of standard backup CBFs \cite{gurriet2020scalable,chen2021backup}. After \(N\) steps, every vehicle has stopped, and the standstill states in \(\mathcal S\) form the terminal invariant set. This set needs no synthesis because the backup reaches it within a known number of steps.

The prediction depends on \(\bm u\) only through the speed and steering angle at which the backup starts, because \(\bm\xi_{k+1|k}\) does not depend on \(\bm u\) (black outline in \cref{fig:backup}a). For example, a control that accelerates raises the speed at which the backup starts, so the vehicle travels farther before it stops (blue above gray in \cref{fig:backup}b). A control that steers away from the route starts the backup with an outward steering angle, so the predicted path swings outward before the backup returns it to the centerline (blue in \cref{fig:backup}a). A condition on this prediction thus constrains only the present control. It needs no optimization over future controls, which keeps the filter computationally efficient.

\subsection{Backup Value and the Centralized Condition}\label{sec:centralized}

For a constraint \(e \in \mathcal E\), the backup value \(B_e: \mathcal X^n \to \mathbb R\) is the smallest margin it attains along the backup prediction from \(\bm x_k\) (minimum of the gray curve in \cref{fig:backup}c):
\begin{equation}
B_e(\bm x_k) = \min_{t \ge 1} h_e\big(\bm\xi(\bm\varphi_t(\bm x_k))\big) = \min_{1 \le t \le N} h_e\big(\bar{\bm\xi}_{k+t|k}\big),\label{eq:backup-value}
\end{equation}
 where the second equality follows from \cref{lem:stopping}. Since the minimum starts at the next pose, \(B_e(\bm x) \ge 0\) does not imply that the current pose is safe. We call \(\bm x\) recoverable if \(B_e(\bm x) \ge 0,\ \forall e \in \mathcal E\). The \emph{certified set} \(\mathcal C\) contains the recoverable states whose current pose is also safe:
\begin{equation}
\mathcal C = \{\bm x \in \mathcal S : B_e(\bm x) \ge 0,\ \forall e \in \mathcal E\}.\label{eq:certified}
\end{equation}
Every state of \(\mathcal C\) is safe now (\(\bm x \in \mathcal S\)) and stays safe forever under the backup (\(B_e(\bm x) \ge 0\)). By \cref{lem:stopping},
\begin{equation}
B_e\big(\bm F(\bm x_k, \bm u)\big) = \min_{2 \le t \le N} h_e\big(\bm\xi_{k+t|k}(\bm u)\big),\label{eq:value-step}
\end{equation}
 so the backup value one step ahead is a finite minimum over poses that the current control determines: once \(\bm u\) is applied, the backup prediction made at step \(k+1\) coincides with the candidate prediction made at step \(k\), \(\bar{\bm\xi}_{k+t|k+1} = \bm\xi_{k+t|k}(\bm u)\) for \(t \ge 2\). Moreover, \(h_e(\bm\xi_{k+1|k}) \ge B_e(\bm x_k)\), since \(\bm\xi_{k+1|k} = \bar{\bm\xi}_{k+1|k}\).

The centralized BH-CBF condition requires
\begin{equation}
h_e\big(\bm\xi_{k+t|k}(\bm u)\big) \ge (1-\lambda)\, B_e(\bm x_k), \quad t = 2, \dots, N,\label{eq:bh-cbf}
\end{equation}
for every constraint \(e\). By \eqref{eq:value-step}, \eqref{eq:bh-cbf} is equivalent to \(B_e(\bm x_{k+1}) \ge (1-\lambda)\, B_e(\bm x_k)\): a discrete-time CBF condition with decay rate \(\lambda\) on the backup value (circled in \cref{fig:backup}c: the minimum of the blue curve over \(t \ge 2\) must stay above the red line).

\begin{definition}[Centralized BH-CBF Filter]\label{def:cen-filter}
The centralized BH-CBF filter \(\bm\Gamma: \mathcal X^n \times \mathbb R^{2n} \to \mathbb R^{2n}\) maps the joint state \(\bm x_k\) and the nominal controls \(\bm u_{\mathrm{nom},k}\) at step \(k\) to an admissible control \(\bm u \in \mathcal U(\bm x_k)\) closest to \(\bm u_{\mathrm{nom},k}\) that satisfies \eqref{eq:bh-cbf} for every constraint:
\begin{equation}
\begin{aligned}
\bm\Gamma(\bm x_k, \bm u_{\mathrm{nom},k}) \in \arg\min_{\bm u \in \mathcal U(\bm x_k)}\ & \|\bm u - \bm u_{\mathrm{nom},k}\|_{\bm W}^2\\
\text{s.t.}\ & \eqref{eq:bh-cbf},\ \forall e \in \mathcal E,
\end{aligned}\label{eq:filter}
\end{equation}
where \(\|\cdot\|_{\bm W}\) is the norm weighted by a positive definite matrix \(\bm W \in \mathbb R^{2n \times 2n}\).
\end{definition}
We call \eqref{eq:filter} the centralized filter problem, and \cref{thm:centralized} states its guarantees.

\begin{theorem}[Centralized Safety]\label{thm:centralized}
Assume \ref{as:plant}. Then:

\begin{enumerate}[label=(\roman*), ref=(\roman*)]
\item\label{thm:centralized-feas}
  if \(\bm x_k \in \mathcal C\), the joint backup control \(\bm u = \bm\kappa(\bm x_k)\) is feasible for \eqref{eq:filter};
\item\label{thm:centralized-inv}
  if \(\bm x_{k_0} \in \mathcal C\) at an initial step \(k_0\) and the applied control is feasible for \eqref{eq:filter} at every step \(k \ge k_0\), then \(\bm x_k \in \mathcal C\) for all \(k \ge k_0\), so by \ref{thm:centralized-feas} a feasible control exists at every step. Hence no lane-boundary or inter-vehicle collision occurs at any sampling instant \(k \ge k_0\).
\end{enumerate}
\end{theorem}

\emph{Proof.} (i) \(\bm F(\bm x_k, \bm\kappa(\bm x_k)) = \bm\varphi_1(\bm x_k)\), so by \ref{k:decoupled} \(\bm\xi_{k+t|k}(\bm\kappa(\bm x_k)) = \bar{\bm\xi}_{k+t|k}\), and \(h_e(\bar{\bm\xi}_{k+t|k}) \ge B_e(\bm x_k) \ge (1-\lambda) B_e(\bm x_k)\) because \(B_e(\bm x_k) \ge 0\). (ii) Let \(\bm x_k \in \mathcal C\), and let the applied control \(\bm u\) be feasible for \eqref{eq:filter}. By \eqref{eq:value-step}, the next state \(\bm x_{k+1} = \bm F(\bm x_k, \bm u)\) satisfies \(B_e(\bm x_{k+1}) \ge (1-\lambda) B_e(\bm x_k) \ge 0,\ \forall e \in \mathcal E\). Moreover, \(h_e(\bm\xi(\bm x_{k+1})) = h_e(\bm\xi_{k+1|k}) \ge B_e(\bm x_k) \ge 0\), so \(\bm x_{k+1} \in \mathcal C\), and \eqref{eq:filter} is feasible at step \(k+1\) by (i). Induction over \(k \ge k_0\) gives \(\bm x_k \in \mathcal C\) for all \(k \ge k_0\), and nonnegative margins guarantee the clearances (\cref{sec:margins}). \(\square\)

\Cref{thm:centralized}\ref{thm:centralized-feas} states that every \(B_e\) is a valid discrete-time CBF under the control bounds. Moreover, the backup satisfies the conditions of all constraints at once, so they never conflict.

\begin{remark}[Relation to Shielding]\label{rem:backup-cbf}
For \(\lambda = 1\), \eqref{eq:bh-cbf} reduces to \(B_e(\bm F(\bm x_k, \bm u)) \ge 0\), so the next state must be recoverable. Model predictive shielding checks the same condition with the same model and backup \cite{bastani2021safea,zhang2019mamps}. It replaces a candidate that fails the check by the backup, whereas our filter \eqref{eq:filter} moves it to the closest control that satisfies \eqref{eq:bh-cbf}. For \(\lambda < 1\), our filter also limits how fast the backup value decreases.
\end{remark}

Two properties of the centralized filter problem shape the rest of this section. First, it needs the joint state and couples the controls of all vehicles in one problem, so its computation time grows with the number of vehicles (addressed in \cref{sec:decentralized}). Second, it is nonconvex because the margins are nonconvex in the controls (addressed in \cref{sec:filter}).

\subsection{Fully Decentralized Filter}\label{sec:decentralized}

\begin{figure}[t]
\centering
% Backup-horizon pair condition and its decentralized split over the prediction; notation as in the text.
% Schematic Euler rollout of two vehicles at a right-angle crossing (figure units: cm), same conventions
% as fig_backup.tex: displacement 0.18 cm per speed unit and step, a_max dt = 0.5, |a_min| dt = 1,
% rectangles of half-length 0.45 and half-width 0.2, straight routes; signed distances and corner
% inequalities computed exactly with eps_veh = 0.1, lambda = 0.7, l_off = 0.5. Panel (a) is drawn at scale 0.75.
% Vehicle i: (0, 0), heading east, v = 5; vehicle j: (1.15, -2.8), heading north, v = 4; N = 7.
% Shares (eq:share): f^{ij} = 1.15, f^{ji} = 2.8, rho^{ij} = 0.333, rho^{ji} = 0.667.
% Candidates: u^i accelerates (a dt = a_max dt = 0.5), u^j brakes gently (a dt = -0.16); both satisfy
%   their corner inequalities, and j meets its share with equality at t = 3.
% (b) Braking gap dbar - eps (t = 1..7): 1.330 0.790 0.541 0.701 0.866 0.866 0.866, so B^{ij} = 0.541 at
%   t = 3 and (1 - lambda) B^{ij} = 0.162. Budget eta (t = 2..7): 0.628 0.378 0.538 0.703 0.703 0.703,
%   split from the bound upward into rho^{ji} eta (green) and rho^{ij} eta (blue).
%   Candidate pair margin sd(R^i, R^j) - eps: 1.330 0.644 0.828 1.430 1.880 2.150 2.240.
\begin{tikzpicture}[
    font=\footnotesize,
    >=latex,
    pt/.style={circle, fill, inner sep=0pt, minimum size=2.5pt},
    lbl/.style={inner sep=1.5pt},
    bg/.style={fill=white, inner sep=0.8pt},
  ]
  % Rectangle heading east (#1 = H) or north (#1 = V) centered at #2, #3; #4: style
  \newcommand{\rect}[4]{%
    \ifx#1H\draw[#4] (#2-0.45, #3-0.2) rectangle (#2+0.45, #3+0.2);%
    \else\draw[#4] (#2-0.2, #3-0.45) rectangle (#2+0.2, #3+0.45);\fi}

  % ---------------- (a) Predictions of both vehicles ----------------
  \begin{scope}[scale=0.75]
    % Lane boundaries of the two routes, open where the route of j enters (T-junction)
    \draw[thin] (-0.6, 0.5) -- (4.75, 0.5);
    \draw[thin] (-0.6, -0.5) -- (0.65, -0.5) -- (0.65, -3.4) (1.65, -3.4) -- (1.65, -0.5) -- (4.75, -0.5);
    \draw[densely dashed, black!40] (-0.6, 0) -- (4.75, 0) (1.15, -3.4) -- (1.15, 0.5);

    % Backup predictions (gray), rectangles at the critical step t = 3
    \draw[black!45] (0.9, 0) -- (2.7, 0);
    \foreach \x in {1.62, 2.16, 2.52, 2.7} \node[pt, black!45] at (\x, 0) {};
    \draw[black!45] (1.15, -2.08) -- (1.15, -1.0);
    \foreach \y in {-1.54, -1.18, -1.0} \node[pt, black!45] at (1.15, \y) {};
    \rect{H}{2.16}{0}{black!45, thick}
    \rect{V}{1.15}{-1.18}{black!45, thick}

    % Predictions for the candidate controls (blue: vehicle i, green: vehicle j), rectangles at t = 3
    \draw[rwthblue] (0.9, 0) -- (4.14, 0);
    \foreach \x in {1.89, 2.7, 3.33, 3.78, 4.05, 4.14} \node[pt, rwthblue] at (\x, 0) {};
    \draw[rwthgreen] (1.15, -2.08) -- (1.15, -0.395);
    \foreach \y in {-1.389, -0.878, -0.546, -0.395} \node[pt, rwthgreen] at (1.15, \y) {};
    \rect{H}{2.7}{0}{rwthblue, thick}
    \rect{V}{1.15}{-0.878}{rwthgreen, thick}

    % Gap between the backup predictions at t = 3 (closest corners)
    \draw[thick] (1.35, -0.73) -- (1.71, -0.2);
    \node[lbl, anchor=west] (gaptxt) at (2.0, -1.45) {\(\bar d^{ij}_{k+3|k}\)};
    \draw[very thin] (gaptxt.west) -- (1.55, -0.465);

    % Current poses and the predicted poses at k+1, which the controls cannot change
    \rect{H}{0}{0}{thick, fill=rwthblue10}
    \rect{V}{1.15}{-2.8}{thick, fill=rwthgreen10}
    \rect{H}{0.9}{0}{thick}
    \rect{V}{1.15}{-2.08}{thick}
    \node[pt] at (0, 0) {};
    \node[pt] at (1.15, -2.8) {};

    % Labels of vehicle i (above and below its lane)
    \node[lbl, bg, anchor=south] at (-0.15, 0.5) {\(\bm\xi^i_k\)};
    \node[lbl, bg, anchor=south] at (1.0, 0.5) {\(\bm\xi^i_{k+1|k}\)};
    \node[lbl, bg, text=black!55, anchor=south] at (2.35, 0.5) {\(\bar{\bm\xi}^i_{k+3|k}\)};
    \node[lbl, bg, text=rwthblue, anchor=south] at (4.0, 0.5) {\(\bm\xi^i_{k+N|k}\)};
    \node[lbl, bg, text=rwthblue, anchor=north] at (3.0, -0.5) {\(\bm\xi^i_{k+3|k}\)};
    % Labels of vehicle j (left of its lane)
    \node[lbl, text=rwthgreen, anchor=east] at (0.6, -0.72) {\(\bm\xi^j_{k+3|k}\)};
    \node[lbl, text=black!55, anchor=east] at (0.6, -1.3) {\(\bar{\bm\xi}^j_{k+3|k}\)};
    \node[lbl, anchor=east] at (0.6, -2.08) {\(\bm\xi^j_{k+1|k}\)};
    \node[lbl, anchor=east] at (0.6, -2.8) {\(\bm\xi^j_k\)};
  \end{scope}
  \node at (1.55, -3.05) {(a)};

  % ---------------- (b) Pair margin and budget along the predictions ----------------
  \begin{scope}[xshift=4.1cm, yshift=-2.45cm, x=0.33cm, y=1.25cm]
    % Budget split between the vehicles (from the bound upward: share of j, share of i)
    \fill[rwthgreen25] (2, 0.162) -- (7, 0.162) -- (7, 0.631) -- (5, 0.631) -- (4, 0.521) -- (3, 0.414)
      -- (2, 0.581) -- cycle;
    \fill[rwthblue25] (2, 0.581) -- (3, 0.414) -- (4, 0.521) -- (5, 0.631) -- (7, 0.631) -- (7, 0.866)
      -- (5, 0.866) -- (4, 0.701) -- (3, 0.541) -- (2, 0.79) -- cycle;
    \fill[rwthred10] (0, 0) rectangle (7, 0.162);
    \draw[->] (0, 0) -- (8.1, 0) node[lbl, below] {\(t\)};
    \draw[->] (0, 0) -- (0, 1.75) node[lbl, right] {pair margin};
    \foreach \t in {0, 1, 2, 3} \draw (\t, 0) -- (\t, -0.06) node[lbl, below] {\(\t\)};
    \draw (7, 0) -- (7, -0.06) node[lbl, below] {\(N\)};
    % Bound of the pair condition
    \draw[rwthred] (0, 0.162) -- (7, 0.162) node[lbl, right] {\((1-\lambda) B^{ij}\)};
    \node[lbl, text=rwthblue, anchor=west] at (7, 0.77) {\(\rho^{ij}\eta^{ij}_{k+t|k}\)};
    \node[lbl, text=rwthgreen, anchor=west] at (7, 0.46) {\(\rho^{ji}\eta^{ij}_{k+t|k}\)};
    % Braking gap (gray), minimum B^{ij} at t = 3
    \draw[black!45] (1, 1.33) -- (2, 0.79) -- (3, 0.541) -- (4, 0.701) -- (5, 0.866) -- (7, 0.866);
    \foreach \t/\d in {1/1.33, 2/0.79, 3/0.541, 4/0.701, 5/0.866, 6/0.866, 7/0.866} \node[pt, black!45] at (\t, \d) {};
    \node[lbl, text=black!55, anchor=south west] at (3.95, 0.93) {\(\bar d^{ij}_{k+t|k} - \varepsilon_{\mathrm{veh}}\)};
    \node[lbl, anchor=east] (Btxt) at (1.75, 0.4) {\(B^{ij}\)};
    \draw[very thin] (Btxt.east) -- (2.93, 0.53);
    % Candidate pair margin (black), leaves the plot after t = 4
    \begin{scope}
      \clip (0, 0) rectangle (7.8, 1.7);
      \draw (1, 1.33) -- (2, 0.644) -- (3, 0.828) -- (4, 1.43) -- (5, 1.88);
    \end{scope}
    \foreach \t/\d in {1/1.33, 2/0.644, 3/0.828, 4/1.43} \node[pt] at (\t, \d) {};
  \end{scope}
  \node at (5.4, -3.05) {(b)};
\end{tikzpicture}
\caption{Candidate predictions of vehicles \(i\) (blue) and \(j\) (green) and their backup predictions (gray) at step \(k\) over \(t\) (schematic). (a) Poses at a crossing; the backup predictions come closest at \(t = 3\). (b) Pair margin \(h^{ij}\) of the backup predictions (gray) and of the candidate predictions (black). The budget between the gray curve and \((1-\lambda) B^{ij}(\bm x_k)\) is split into the shares of \(i\) (blue fill) and \(j\) (green fill).}
\label{fig:pair}
\end{figure}
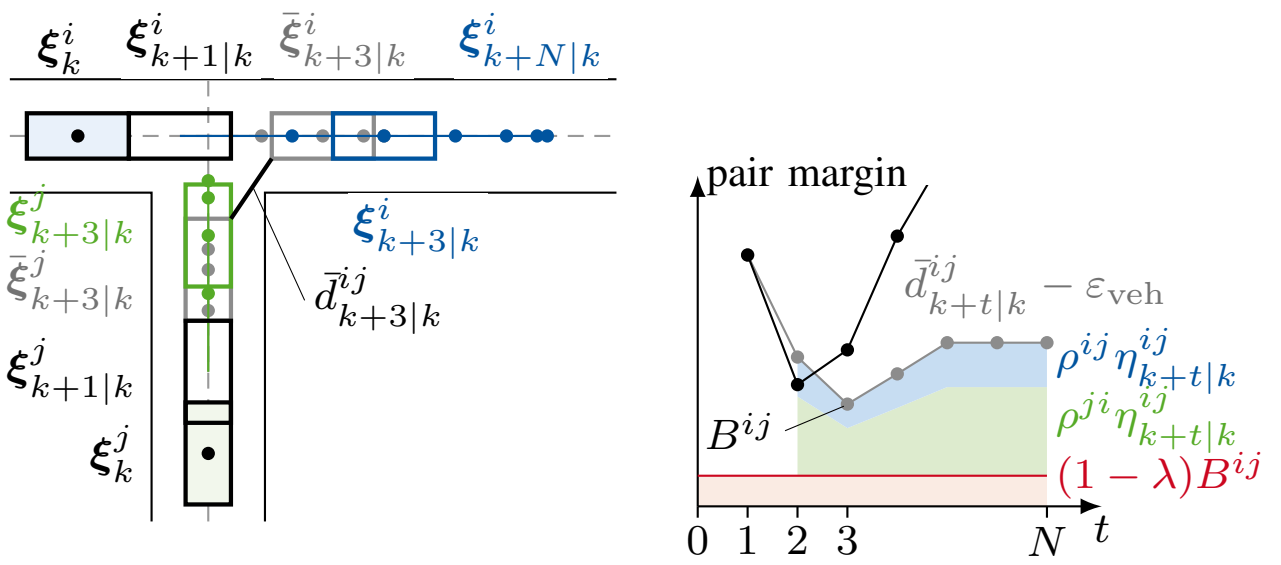

\begin{figure}[t]
\centering
% Decentralized corner inequalities at one prediction step t; notation as in the text.
% Rectangles of half-length 0.7 and half-width 0.3 (figure units: cm); the scene is rotated so that the
% separating normal n^{ij} points up.
% Backup predictions: R^j at (1.6, 0), heading 0; R^i at (1.249, 1.995), heading -8 deg, closest corner
%   (1.9, 1.6) above the top edge of R^j, so sd = gap along n = 1.3 and the separating normal is exactly vertical.
% Budget split: rho^{ji} eta = 0.35, eps_veh + (1 - lambda) B^{ij} = 0.45, rho^{ij} eta = 0.5.
% Candidates (blue: vehicle i, green: vehicle j): R^i at (2.195, 1.621), heading -20 deg, lowest corner
%   (2.75, 1.1) on its limit (full share);
%   R^j at (2.185, 0.178), heading 6 deg, highest corner (2.85, 0.55) below its limit (part of its share).
\begin{tikzpicture}[
    font=\footnotesize,
    >=latex,
    pt/.style={circle, fill, inner sep=0pt, minimum size=3.5pt},
    lbl/.style={inner sep=1.5pt},
    bg/.style={fill=white, inner sep=0.8pt},
    dim/.style={<->, thin},
  ]
  % Gap between the backup predictions along the separating normal and its split
  \fill[rwthgreen25] (-0.1, 0.3) rectangle (4.1, 0.65);
  \fill[rwthred10] (-0.1, 0.65) rectangle (4.1, 1.1);
  \fill[rwthblue25] (-0.1, 1.1) rectangle (4.1, 1.6);
  \draw[rwthred, thin] (-0.1, 0.65) -- (4.1, 0.65) (-0.1, 1.1) -- (4.1, 1.1);
  \draw[densely dashed, black!45] (-0.1, 0.3) -- (4.1, 0.3) (-0.1, 1.6) -- (4.1, 1.6);
  \draw[dim] (4.0, 0.3) -- node[lbl, right, text=rwthgreen] {\(\rho^{ji} \eta^{ij}_{k+t|k}\)} (4.0, 0.65);
  \draw[dim] (4.0, 0.65) -- node[lbl, right, text=rwthred] {\(\varepsilon_{\mathrm{veh}} + (1-\lambda) B^{ij}(\bm x_k)\)} (4.0, 1.1);
  \draw[dim] (4.0, 1.1) -- node[lbl, right, text=rwthblue] {\(\rho^{ij} \eta^{ij}_{k+t|k}\)} (4.0, 1.6);
  \draw[dim] (-0.1, 0.3) -- node[lbl, left] {\(\bar d^{ij}_{k+t|k}\)} (-0.1, 1.6);
  \draw[->, thick] (0.3, 1.17) -- (0.3, 1.55) node[lbl, right, pos=0.5] {\(\bm n^{ij}_{k+t|k}\)};

  % Backup predictions (gray)
  \draw[black!45, thick] (2.3, 0.3) -- (2.3, -0.3) -- (0.9, -0.3) -- (0.9, 0.3) -- cycle;
  \draw[black!45, thick] (1.984, 2.194) -- (1.9, 1.6) -- (0.514, 1.795) -- (0.597, 2.389) -- cycle;
  \node[pt, black!45] at (1.9, 1.6) {};
  \node[lbl, text=black!55, anchor=east] at (0.85, 0) {\(\bar{\mathcal R}^j_{k+t|k}\)};
  \node[lbl, text=black!55, anchor=east] at (0.5, 2.1) {\(\bar{\mathcal R}^i_{k+t|k}\)};

  % Candidate predictions (blue: vehicle i, green: vehicle j)
  \draw[rwthblue, thick] (2.955, 1.664) -- (2.75, 1.1) -- (1.434, 1.579) -- (1.64, 2.143) -- cycle;
  \draw[rwthgreen, thick] (2.85, 0.55) -- (2.913, -0.047) -- (1.52, -0.193) -- (1.458, 0.404) -- cycle;
  \node[pt, rwthblue] at (2.75, 1.1) {};
  \node[lbl, bg, text=rwthblue, anchor=west] at (3.0, 1.85) {\(\mathcal R^i_{k+t|k}\)};
  \node[lbl, bg, text=rwthgreen, anchor=west] at (3.0, 0.05) {\(\mathcal R^j_{k+t|k}\)};
\end{tikzpicture}
\caption{Decentralized corner inequalities \eqref{eq:dec-corner} at step \(k\) for one prediction step \(t\) (schematic), drawn with the separating normal \(\bm n^{ij}_{k+t|k}\). Candidate predictions of vehicles \(i\) (blue) and \(j\) (green) and their backup predictions (gray). Blue and green fills: shares of \(i\) and \(j\), as in \cref{fig:pair}. Gray dot: closest corner of \(\bar{\mathcal R}^i_{k+t|k}\) along \(\bm n^{ij}_{k+t|k}\); blue dot: a corner of \(\mathcal R^i_{k+t|k}\) that uses the full share \(\rho^{ij}\eta^{ij}_{k+t|k}\) of the budget.}
\label{fig:decentralized}
\end{figure}
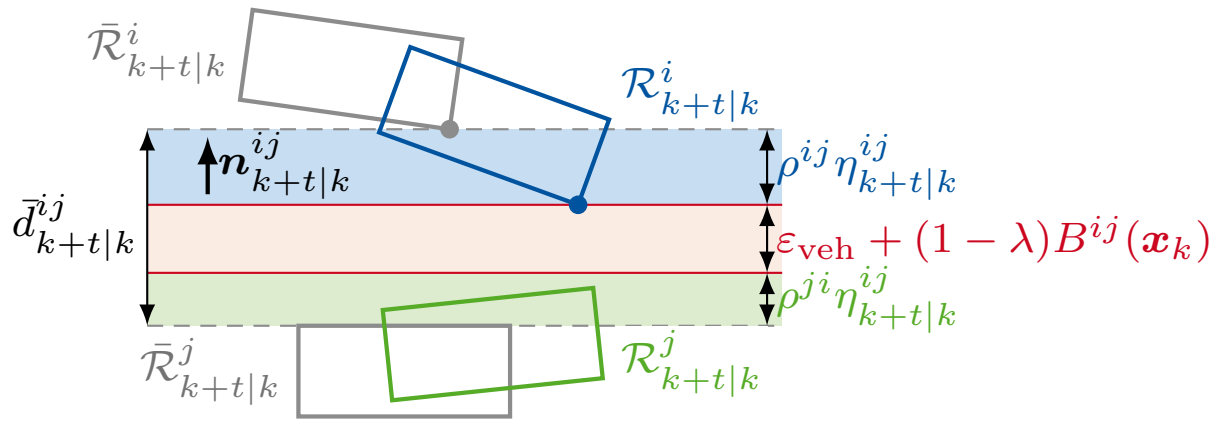

To design a fully decentralized filter, we must split each pair inequality of \eqref{eq:bh-cbf} into one inequality per vehicle, because it couples the controls of both vehicles. Each vehicle then takes responsibility for a share of the allowed decrease of the pair margin. The decentralized backup CBF \cite{chen2021guaranteed} uses a half split, in which each vehicle covers half of the allowed decrease of a smooth pair barrier. This split is exact in continuous time by the chain rule. Over a finite step, the change of the pair margin when both vehicles deviate from their backup predictions differs from the sum of the one-sided changes by a mixed term. This term is caused by the nonlinearity of the distance and by the separating directions of rectangles, which turn with the headings. The mixed term can be negative, so the two one-sided conditions do not imply the pair condition.

Therefore, we measure both deviations along a direction fixed in advance. Consider a pair of vehicles \(i\) and \(j\). Each of them computes both backup predictions (its own and that of the other vehicle), \(\bar{\bm\xi}^i_{k+t|k} = \bm\xi(\bm\varphi^i_t(\bm x^i_k))\) and \(\bar{\bm\xi}^j_{k+t|k}\), from the states and routes that both broadcast at every step (\ref{as:coop}, \ref{k:decoupled}). Let \(\bar{\mathcal R}^i_{k+t|k} = \mathcal R(\bar{\bm\xi}^i_{k+t|k})\), and let \(\bar d^{ij}_{k+t|k} = \mathrm{sd}(\bar{\mathcal R}^i_{k+t|k}, \bar{\mathcal R}^j_{k+t|k})\) be their signed distance and \(\bm n^{ij}_{k+t|k}\) their separating normal. The backup value \(B^{ij}\) of the pair is \(B_e\) in \eqref{eq:backup-value} for the pair constraint, \(B^{ij}(\bm x_k) = \min_{1 \le t \le N} h^{ij}(\bar{\bm\xi}^i_{k+t|k}, \bar{\bm\xi}^j_{k+t|k})\). Each term is the pair margin of the backup predictions, \(h^{ij}(\bar{\bm\xi}^i_{k+t|k}, \bar{\bm\xi}^j_{k+t|k}) = \bar d^{ij}_{k+t|k} - \varepsilon_{\mathrm{veh}}\) (gray curve in \cref{fig:pair}b; \(B^{ij}\) marks its minimum). Let \(\rho^{ij}, \rho^{ji} \ge 0\) denote the responsibility shares of \(i\) and \(j\). The shares can be chosen freely as long as \(\rho^{ij} + \rho^{ji} = 1\). We use the heuristic split
\begin{equation}
\rho^{ij} = (f^{ij} + \ell_{\mathrm{off}})/(f^{ij} + f^{ji} + 2\ell_{\mathrm{off}}),\label{eq:share}
\end{equation}
where \(f^{ij} = \max(0, \langle \bm e^i_{\mathrm{lon}}, \bm p^j - \bm p^i \rangle)\) is how far \(j\) lies ahead of \(i\) and the offset \(\ell_{\mathrm{off}} > 0\) splits evenly when neither vehicle lies ahead of the other. Responsibility thus lies with the vehicle that approaches the other, like the rear vehicle in car following (in \cref{fig:pair}, \(j\) approaches the route of \(i\) and receives the larger share, see green fill). A half split would instead waste half of the budget on a vehicle that moves away and cannot use it. For each of its corners \(c = 1, \dots, 4\) and \(t = 2, \dots, N\), vehicle \(i\) replaces the pair inequalities of the centralized condition \eqref{eq:bh-cbf} by the decentralized condition
\begin{equation}
\begin{aligned}
&\big\langle \bm n^{ij}_{k+t|k}, \bm q_c(\bm\xi^i_{k+t|k}(\bm u^i)) \big\rangle - \min_{c'} \big\langle \bm n^{ij}_{k+t|k}, \bm q_{c'}(\bar{\bm\xi}^i_{k+t|k}) \big\rangle\\
& \ge -\rho^{ij}\, \eta^{ij}_{k+t|k},\ \forall c \in \{1, \dots, 4\},\ t \in \{2, \dots, N\},\\
&\text{with } \eta^{ij}_{k+t|k} = \bar d^{ij}_{k+t|k} - \varepsilon_{\mathrm{veh}} - (1-\lambda) B^{ij}(\bm x_k).
\end{aligned}\label{eq:dec-corner}
\end{equation}
Let \(\mathcal R^i_{k+t|k} = \mathcal R(\bm\xi^i_{k+t|k}(\bm u^i))\) and \(\mathcal R^j_{k+t|k} = \mathcal R(\bm\xi^j_{k+t|k}(\bm u^j))\) be the candidate rectangles of vehicles \(i\) and \(j\). The left side of \eqref{eq:dec-corner} measures how far corner \(c\) of \(\mathcal R^i_{k+t|k}\) lies beyond the closest corner of \(\bar{\mathcal R}^i_{k+t|k}\), along the direction away from \(j\); the right side is its share \(\rho^{ij}\) of the displacement budget \(\eta^{ij}_{k+t|k}\) (\cref{fig:decentralized}: every corner of the blue rectangle of \(i\) stays at most \(\rho^{ij}\eta^{ij}_{k+t|k}\) below the gray dot). Only the side of the rectangle that faces \(j\) matters, so corners on the far side may move as long as they stay behind that closest corner, and \eqref{eq:dec-corner} does not lock the heading (the blue rectangle of \(i\) in \cref{fig:decentralized} has turned). Vehicle \(j\) imposes the mirrored inequalities with its own separating normal \(\bm n^{ji}_{k+t|k} = -\bm n^{ij}_{k+t|k}\), the same budget \(\eta^{ji}_{k+t|k} = \eta^{ij}_{k+t|k}\) (all vehicles use the same decay rate \(\lambda\)), and its share \(\rho^{ji}\). Together, the inequalities of both vehicles keep the red band in \cref{fig:decentralized} free: of the gap \(\bar d^{ij}_{k+t|k}\) between the backup predictions, each vehicle uses at most its share of the budget, and \(\varepsilon_{\mathrm{veh}} + (1-\lambda) B^{ij}(\bm x_k)\) remains. We call this split of the budget over the corners of both vehicles the \emph{corner budgets}. The budget varies along the prediction. At the step where the backup predictions come closest, \(\bar d^{ij}_{k+t|k} - \varepsilon_{\mathrm{veh}} = B^{ij}(\bm x_k)\), so the budget shrinks to \(\lambda B^{ij}(\bm x_k)\) (\(t = 3\) in \cref{fig:pair}b, where the blue and green fills are thinnest). At all other steps, it is larger, and the vehicles may deviate further from their backup predictions. Lane constraints involve only the ego vehicle and keep the form \eqref{eq:bh-cbf}. The centralized and the decentralized filter use the same margins, so they share the safe set \(\mathcal S\) and the certified set \(\mathcal C\) in \eqref{eq:certified}. 

Each vehicle needs to coordinate only with the vehicles that it can reach before both have stopped. Under any admissible first control, the center of vehicle \(i\) with speed \(v\) travels at most \(D^i(v) = v\Delta t + 2 \hat v \Delta t + \hat v^2 / (2|a^i_{\min}|)\) along the prediction \(\bm\xi^i_{k+1|k}, \dots, \bm\xi^i_{k+N|k}\), where \(\hat v = v + a^i_{\max}\Delta t\) bounds the speed after one step and the last two terms bound the Euler braking distance from \(\hat v\). We call \(D^i: [0, v_{\max}] \to \mathbb R_{\ge 0}\) the \emph{reach} of vehicle \(i\). The neighbor set \(\mathcal N^i \subseteq \mathcal I \setminus \{i\}\) of vehicle \(i\) contains every vehicle \(j\) with \(\|\bm p^i - \bm p^j\| \le D^i(v^i) + D^j(v^j) + \ell^i_{\mathrm c} + \ell^j_{\mathrm c} + \varepsilon_{\mathrm{veh}}\), where \(\ell^i_{\mathrm c} = ((\ell^i_{\mathrm{lon}})^2 + (\ell^i_{\mathrm{lat}})^2)^{1/2}\) is the circumradius of the rectangle of vehicle \(i\). For every other pair, the triangle inequality keeps the predicted rectangles farther apart than \(\varepsilon_{\mathrm{veh}}\) for all admissible controls, so \(B^{ij}(\bm x_{k+1}) \ge 0\) holds without any inequality. The neighbor relation is symmetric, so both vehicles of a pair agree on whether to impose \eqref{eq:dec-corner}.

\begin{definition}[Decentralized BH-CBF Filter]\label{def:dec-filter}
Our decentralized BH-CBF filter \(\bm\Gamma^i: \mathcal X^n \times \mathbb R^2 \to \mathbb R^2\) of vehicle \(i\) maps the joint state \(\bm x_k\) and the nominal control \(\bm u^i_{\mathrm{nom},k}\) of vehicle \(i\) at step \(k\) to an admissible control \(\bm u^i \in \mathcal U^i(\bm x^i_k)\) closest to \(\bm u^i_{\mathrm{nom},k}\). This control satisfies the lane inequalities \eqref{eq:bh-cbf} of vehicle \(i\) and its corner inequalities \eqref{eq:dec-corner} with every neighbor \(j \in \mathcal N^i\):
\begin{equation}
\begin{aligned}
\bm\Gamma^i(\bm x_k, \bm u^i_{\mathrm{nom},k}) \in \arg\min_{\bm u^i \in \mathcal U^i(\bm x^i_k)}\ & \|\bm u^i - \bm u^i_{\mathrm{nom},k}\|_{\bm W^i}^2\\
\text{s.t.}\ & \eqref{eq:bh-cbf},\ \forall e \in \mathcal E^i_{\mathrm L} \cup \mathcal E^i_{\mathrm R},\\
& \eqref{eq:dec-corner},\ \forall j \in \mathcal N^i,
\end{aligned}\label{eq:dec-filter}
\end{equation}
where \(\bm W^i \in \mathbb R^{2 \times 2}\) is a positive definite weight matrix.
\end{definition}
We call \eqref{eq:dec-filter} the filter problem of vehicle \(i\), and \cref{thm:decentralized} states its guarantees.

\begin{theorem}[Decentralized Safety]\label{thm:decentralized}
Assume \ref{as:plant} and \ref{as:coop}. Then:

\begin{enumerate}[label=(\roman*), ref=(\roman*)]
\item\label{thm:decentralized-feas}
  if \(\bm x_k \in \mathcal C\), the backup control \(\bm u^i = \bm\kappa^i(\bm x^i_k)\) of each vehicle \(i\) is feasible for its filter problem \eqref{eq:dec-filter}, which involves no control of another vehicle;
\item\label{thm:decentralized-inv}
  if \(\bm x_{k_0} \in \mathcal C\) at an initial step \(k_0\) and every vehicle applies a feasible control of its filter problem at every step \(k \ge k_0\), then \(\bm x_k \in \mathcal C\) for all \(k \ge k_0\), so by \ref{thm:decentralized-feas} every filter problem is feasible at every step. Hence no lane-boundary or inter-vehicle collision occurs at any sampling instant \(k \ge k_0\).
\end{enumerate}
\end{theorem}

\emph{Proof.} Let \(\bm x_k \in \mathcal C\). For each pair with \(j \in \mathcal N^i\), both vehicles compute every quantity in \eqref{eq:dec-corner} from the same broadcast states, so they agree on it. Since \(B^{ij}(\bm x_k) \ge 0\), the backup predictions are disjoint (\(\bar d^{ij}_{k+t|k} \ge \varepsilon_{\mathrm{veh}} > 0\)), so the separating normal is unique, and vehicle \(j\) obtains exactly \(-\bm n^{ij}_{k+t|k}\). Any horizon that satisfies \cref{lem:stopping} for both vehicles gives the same values. Hence the inequalities of \(j\) are those of \(i\) mirrored, with the separating normal \(-\bm n^{ij}_{k+t|k}\) and the same budget. Since the minimum in \(B^{ij}\) includes every \(t \ge 1\), \(\bar d^{ij}_{k+t|k} - \varepsilon_{\mathrm{veh}} \ge B^{ij}\), hence \(\eta^{ij}_{k+t|k} \ge \lambda B^{ij} \ge 0\). (i) With its own backup, \(\bm\xi^i_{k+t|k}(\bm\kappa^i(\bm x^i_k)) = \bar{\bm\xi}^i_{k+t|k}\), so the left side of \eqref{eq:dec-corner} is \(\langle \bm n^{ij}_{k+t|k}, \bm q_c(\bar{\bm\xi}^i_{k+t|k}) \rangle - \min_{c'} \langle \bm n^{ij}_{k+t|k}, \bm q_{c'}(\bar{\bm\xi}^i_{k+t|k}) \rangle \ge 0 \ge -\rho^{ij}\eta^{ij}_{k+t|k}\), and the lane inequalities hold as in \cref{thm:centralized}\ref{thm:centralized-feas}. (ii) Let every vehicle apply a feasible control of its filter problem at step \(k\). Taking the minimum over the corners \(c\) in the inequalities of \(i\) gives \(\min_c \langle \bm n^{ij}_{k+t|k}, \bm q_c(\mathcal R^i_{k+t|k})\rangle \ge \min_c \langle \bm n^{ij}_{k+t|k}, \bm q_c(\bar{\mathcal R}^i_{k+t|k})\rangle - \rho^{ij}\eta^{ij}_{k+t|k}\), and the inequalities of \(j\) with \(\bm n^{ji}_{k+t|k} = -\bm n^{ij}_{k+t|k}\) give \(\max_c \langle \bm n^{ij}_{k+t|k}, \bm q_c(\mathcal R^j_{k+t|k})\rangle \le \max_c \langle \bm n^{ij}_{k+t|k}, \bm q_c(\bar{\mathcal R}^j_{k+t|k})\rangle + \rho^{ji}\eta^{ij}_{k+t|k}\). Subtracting gives \(\mathrm{gap}_{\bm n^{ij}_{k+t|k}}(\mathcal R^i_{k+t|k}, \mathcal R^j_{k+t|k}) \ge \mathrm{gap}_{\bm n^{ij}_{k+t|k}}(\bar{\mathcal R}^i_{k+t|k}, \bar{\mathcal R}^j_{k+t|k}) - \eta^{ij}_{k+t|k} = \bar d^{ij}_{k+t|k} - \eta^{ij}_{k+t|k} = \varepsilon_{\mathrm{veh}} + (1-\lambda) B^{ij}(\bm x_k)\) by \eqref{eq:sd} at the separating normal. Then \(\mathrm{sd}(\mathcal R^i_{k+t|k}, \mathcal R^j_{k+t|k}) - \varepsilon_{\mathrm{veh}} \ge (1-\lambda) B^{ij}(\bm x_k)\) by \eqref{eq:sd}. By \eqref{eq:value-step} with \(\bm x_{k+1} = \bm F(\bm x_k, \bm u)\) and by \ref{k:decoupled}, the minimum over \(t = 2, \dots, N\) equals \(B^{ij}(\bm x_{k+1})\), so \(B^{ij}(\bm x_{k+1}) \ge (1-\lambda) B^{ij}(\bm x_k) \ge 0\). Pairs outside the neighbor sets keep \(B^{ij}(\bm x_{k+1}) \ge 0\) by the choice of \(\mathcal N^i\), and the lane inequalities give \(B_e(\bm x_{k+1}) \ge (1-\lambda) B_e(\bm x_k) \ge 0\) as in the proof of \cref{thm:centralized}\ref{thm:centralized-inv}. At \(k+1\), every pair satisfies \(\mathrm{sd}(\mathcal R(\bm\xi^i_{k+1}), \mathcal R(\bm\xi^j_{k+1})) - \varepsilon_{\mathrm{veh}} = \bar d^{ij}_{k+1|k} - \varepsilon_{\mathrm{veh}} \ge B^{ij}(\bm x_k) \ge 0\), and every lane constraint satisfies \(h_e(\bm\xi_{k+1|k}) \ge B_e(\bm x_k) \ge 0\), so \(\bm x_{k+1} \in \mathcal S\). With the bounds on \(B^{ij}\) and \(B_e\) above, \(\bm x_{k+1} \in \mathcal C\), and all filter problems are feasible at step \(k+1\) by (i). Induction over \(k \ge k_0\) gives \(\bm x_k \in \mathcal C\) for all \(k \ge k_0\), and nonnegative margins guarantee the clearances (\cref{sec:margins}). \(\square\)

\Cref{thm:decentralized}\ref{thm:decentralized-feas} is a decentralized validity: the inequalities of vehicle \(i\) are valid with respect to its own control alone, whatever the other vehicles apply. Each step needs one broadcast of states and routes (\ref{as:coop}) and no exchange of controls, since every quantity in \eqref{eq:dec-corner} is fixed by \(\bm x_k\). The theorem holds for any shares with \(\rho^{ij} + \rho^{ji} = 1\), so the split \eqref{eq:share} affects only how much each vehicle may deviate from its backup, not safety.

The proof of \cref{thm:decentralized}\ref{thm:decentralized-inv} shows that if both vehicles of a neighbor pair satisfy \eqref{eq:dec-corner}, the pair satisfies \eqref{eq:bh-cbf}. The converse does not hold, since \eqref{eq:dec-corner} fixes the direction \(\bm n^{ij}_{k+t|k}\) and the split of the budget. The decentralized filter thus rejects joint controls in which one vehicle uses more than its share while the other uses less, which is the price of exchanging no controls.

\begin{remark}[Relation to Contingency Model-Based Control]\label{rem:cmc}
For identical disks with the closest point in place of the closest corner, \(\lambda = 1\), and \(\rho^{ij} = 1/2\), the inequalities \eqref{eq:dec-corner} reduce to the constraint of contingency model-based control \cite{schildbach2026contingency}. This constraint keeps each spherical agent behind the plane that bisects the braking positions of both agents. Our corner budgets extend it to rectangular vehicles whose heading changes, to a budget that decays with \(\lambda\), and to arbitrary complementary shares.
\end{remark}

\subsection{Certified Solution of the Filter Problems}\label{sec:filter}

A solver for the nonconvex problems \eqref{eq:filter} and \eqref{eq:dec-filter} may return no solution or a control that violates their inequalities. However, by \cref{thm:centralized,thm:decentralized}, any feasible control keeps the guarantees, so it suffices to find a candidate control close to the nominal one and to certify it exactly. Our filter therefore computes the candidate by a QP and accepts it only if it satisfies every inequality of the filter problem exactly. If no candidate passes this exact check, the filter falls back to the backup, which always passes. We call the result the certified filter.

The QP keeps the objective of the filter problem and replaces each inequality by its linearization at the nominal controls, computed by finite differences (step \(10^{-4}\)). Linearization errors can make this QP infeasible even where the filter problem is feasible. Therefore, we relax the linearized inequalities of each constraint by a slack variable that is penalized in the objective. The slacks affect only the candidate, since the exact check uses no slack. If the check fails, the filter repairs the candidate: it adds the violated inequalities, linearized at the candidate, and solves the QP again. It repeats this at most three times. If the QP has no solution or the check still fails, the filter switches the vehicles of violated constraints to their backup until the check passes. The centralized filter switches them jointly, and in the decentralized filter, each vehicle can switch only itself. In contrast to an emergency stop that is applied without a guarantee when the QP is infeasible \cite{han2024multiagent}, our fallback is certified (\cref{prop:filter}).

\begin{proposition}[Filter Guarantees]\label{prop:filter}
Let \(\bm x_k \in \mathcal C\). Then the filter terminates after at most one pass of the fallback per vehicle and returns controls that are feasible for the exact filter problem \eqref{eq:filter} or \eqref{eq:dec-filter}. Hence the closed loop satisfies the hypotheses of \cref{thm:centralized,thm:decentralized}.
\end{proposition}

\emph{Proof.} Each pass of the fallback switches at least one vehicle that was not switched before, so there is at most one pass per vehicle. A pair or lane constraint depends only on the controls of its at most two vehicles, by \ref{k:decoupled}. Once all of them are switched, its inequalities hold by \cref{thm:centralized}\ref{thm:centralized-feas} or \cref{thm:decentralized}\ref{thm:decentralized-feas}, so the fallback ends without a violated inequality. For the decentralized filter, every quantity in \eqref{eq:dec-corner} is fixed by \(\bm x_k\), and the fallback changes only its own control, so each vehicle certifies its control independently. \(\square\)

\section{Safeguarding Multi-Agent Reinforcement Learning}\label{sec:safeguard}

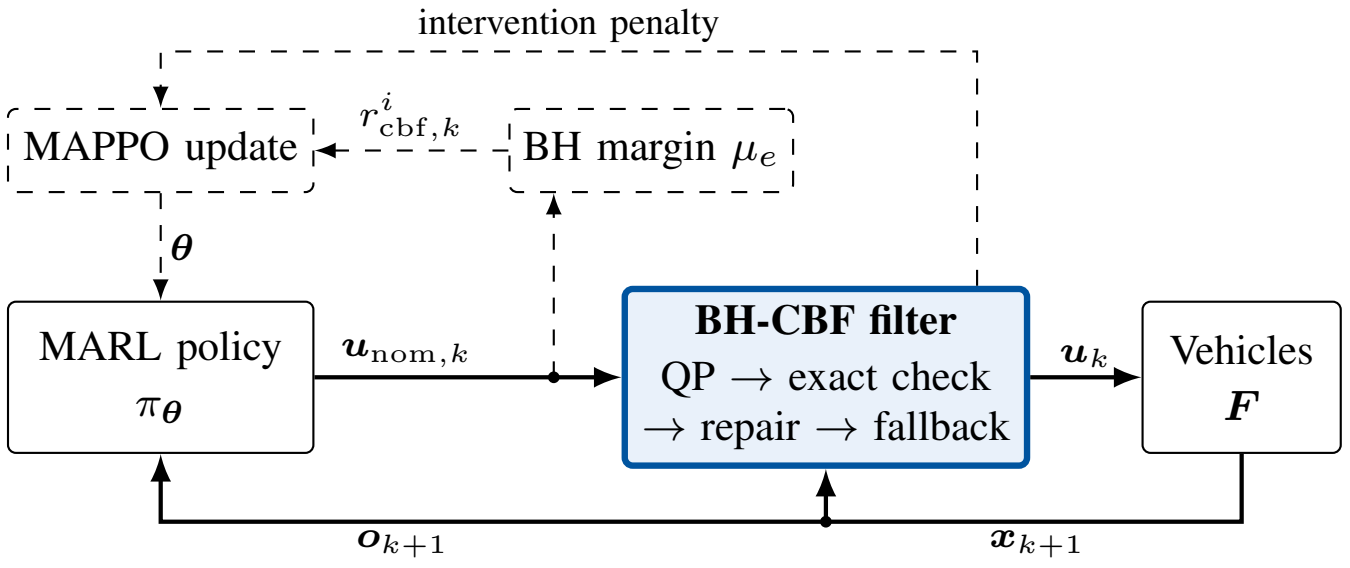
\begin{figure}[t]
\centering
% BH-CBF pipeline: run-time filter with the two training signals (BH-CBF-informed reward, intervention penalty).
\begin{tikzpicture}[
    font=\fontsize{8}{9.5}\selectfont, % fixed sizes (\footnotesize, \scriptsize of a 10pt class) so the layout is the same in all classes
    >=latex,
    box/.style={draw, rounded corners=1.5pt, align=center, inner sep=3pt, minimum height=1.0cm},
    train/.style={draw, dashed, rounded corners=1.5pt, align=center, inner sep=2.5pt, minimum height=0.55cm},
    sig/.style={->, thick},
    tsig/.style={->, dashed},
    lbl/.style={font=\fontsize{7}{8}\selectfont, inner sep=1.5pt},
  ]
  % Run-time loop
  \node[box, minimum width=2.02cm] (pol) at (0, 0) {MARL policy\\[1pt] \(\pi_{\bm\theta}\)};
  \node[box, text width=2.45cm, draw=rwthblue, very thick, fill=rwthblue10] (cbf) at (4.4, 0)
    {\textbf{BH-CBF filter}\\[1pt] QP \(\to\) exact check\\ \(\to\) repair \(\to\) fallback};
  \node[box, text width=1.1cm] (sim) at (7.15, 0) {Vehicles\\[1pt] \(\bm F\)};

  \draw[sig] (pol) -- node[lbl, above, pos=0.3] {\(\bm u_{\mathrm{nom},k}\)} (cbf);
  \draw[sig] (cbf) -- node[lbl, above] {\(\bm u_k\)} (sim);
  \coordinate (fb) at ([yshift=-0.35cm]cbf.south); % below the tallest box
  \draw[sig] (sim.south) -- (sim.south |- fb) -| (pol.south);
  \fill (cbf.south |- fb) circle (1.1pt);
  \draw[sig] (cbf.south |- fb) -- (cbf.south);
  \node[lbl, below] at (5.8, 0 |- fb) {\(\bm x_{k+1}\)};
  \node[lbl, below] at (1.6, 0 |- fb) {\(\bm o_{k+1}\)};

  % Training signals (dashed)
  \node[train, minimum width=2.02cm] (ppo) at (0, 1.5) {MAPPO update};
  \node[train, anchor=west] (mu) at (2.3, 1.5) {BH margin \(\mu_e\)};
  \coordinate (tap) at (2.6, 0);
  \fill (tap) circle (1.1pt);
  \draw[tsig] (tap) -- (tap |- mu.south);
  \draw[tsig] (mu) -- node[lbl, above] {\(r^i_{\mathrm{cbf},k}\)} (ppo);
  \draw[tsig] (cbf.north -| 5.4, 0) -- ++(0, 1.55) -| node[lbl, above, pos=0.25] {intervention penalty} (ppo.north);
  \draw[tsig] (ppo) -- node[lbl, right] {\(\bm\theta\)} (pol);
\end{tikzpicture}
\caption{BH-CBF pipeline. Solid: run-time safety filter. Dashed: training signals.}
\label{fig:pipeline}
\end{figure}

Our BH-CBF safeguards a MARL policy over its life cycle: as a reward during pre-training, as a run-time filter at deployment, and as a filter in the training loop that keeps fine-tuning collision-free, also on unseen maps. \Cref{fig:pipeline} shows how these uses connect. At run time (solid arrows), the policy maps the joint observation \(\bm o_k\) to nominal controls \(\bm u_{\mathrm{nom},k}\). The certified filter of \cref{sec:filter} turns them into the applied controls \(\bm u_k\) and runs fully decentralized (blue box). During training (dashed arrows), two signals enter the update of the policy parameters \(\bm\theta\): a BH-CBF-informed reward \(r^i_{\mathrm{cbf},k}\) on the nominal controls in pre-training and a penalty on the interventions of the filter in fine-tuning. The filter needs only the nominal controls, so it also applies to planners other than MARL policies. By \cref{thm:centralized,thm:decentralized} and \cref{prop:filter}, the filtered system is collision-free at the sampling instants from states in \(\mathcal C\) regardless of the policy, also on maps that the policy has never seen. The policy determines only the efficiency: how often the filter intervenes, how often vehicles fall back to braking, and how quickly they reach their goals.

During pre-training, the policy learns from the margin by which its own action violates the BH-CBF condition \eqref{eq:bh-cbf}. Following our prior work \cite{xu2026safety}, we evaluate the condition at the nominal controls \(\bm u_{\mathrm{nom},k}\) of all vehicles and convert its violation into a reward, without solving a QP. By \eqref{eq:value-step}, the \emph{BH margin} \(\mu_e: \mathcal X^n \times \mathbb R^{2n} \to \mathbb R\) of the nominal action for constraint \(e\) is
\begin{equation*}
\begin{aligned}
\mu_e(\bm x_k, \bm u_{\mathrm{nom},k}) &= B_e\big(\bm F(\bm x_k, \bm u_{\mathrm{nom},k})\big)\\
&\quad - (1-\lambda) \max\big(B_e(\bm x_k), 0\big).
\end{aligned}
\end{equation*}
 For \(B_e(\bm x_k) \ge 0\), \(\mu_e\) is the height of the lowest point of the blue curve for \(t \ge 2\) above the red line in \cref{fig:backup}c, with the blue curve taken for the nominal controls. Then \(\mu_e \ge 0\) if and only if the nominal controls satisfy \eqref{eq:bh-cbf} for \(e\), although the stricter decentralized inequalities \eqref{eq:dec-corner} and the QP of \cref{sec:filter} may still modify the action. Unfiltered training also reaches unrecoverable states. At such states, \(\mu_e = B_e(\bm F(\bm x_k, \bm u_{\mathrm{nom},k}))\), so the penalty persists as long as the next state remains unrecoverable and decreases as the action raises the backup value. The positive part \(\max(B_e, 0)\) prevents a reward for merely reducing a negative value by the factor \(1 - \lambda\). The safety reward of vehicle \(i\) has one channel \(\chi \in \{\mathrm L, \mathrm R, \mathrm V\}\) per constraint type (left side, right side, and vehicle pairs),
\begin{equation}
r^i_{\mathrm{cbf},k} = \sum_{\chi \in \{\mathrm L, \mathrm R, \mathrm V\}} \max\Big(\min\Big(\tfrac{1}{h_{\mathrm{norm},\chi}} \min_{e \in \mathcal E^i_\chi} \mu_e,\ 0\Big),\ -1\Big),\label{eq:cbf-reward}
\end{equation}
 where the constraint sets \(\mathcal E^i_\chi\) are those of \cref{sec:problem}, so a pair violation is charged to both vehicles. Each channel thus penalizes the worst violation, scaled by a normalizer \(h_{\mathrm{norm},\chi} > 0\), and saturates at \(-1\). The total reward is \(r^i_k = r^i_{\mathrm{task},k} + r^i_{\mathrm{cbf},k}\), where the task reward \(r^i_{\mathrm{task},k}\) scores progress, goal reaching, and collisions. Unlike the one-step CBF reward of \cite{xu2026safety}, \eqref{eq:cbf-reward} penalizes an action as soon as the backup can no longer keep the vehicles safe after it. It needs only the prediction along the backup and adds little cost to training. We train the policy with Multi-Agent Proximal Policy Optimization (MAPPO) \cite{yu2022surprising}, which follows centralized training with decentralized execution. We therefore compute the reward from the joint state and action during training, and the trained policy runs decentralized without it.

Fine-tuning adapts a pre-trained policy to a new map. Conventional fine-tuning runs without a filter, so the policy learns to avoid collisions only by colliding. Therefore, we run BH-CBF inside the data collector of MAPPO. The environment executes the filtered control, so the training trajectories are collision-free from states in \(\mathcal C\), while MAPPO stores the policy's sampled action and its log-probability. The filter is a deterministic function of the state and the action, so this is ordinary on-policy learning in a modified environment whose transition applies the filter before the vehicle dynamics \cite{cheng2019endtoend,krasowski2023provably}. However, two problems remain: the filter flattens the policy gradient wherever it maps many actions to the same control, and the policy is trained with the filter but may be deployed without it. An intervention penalty addresses both problems. It reduces the reward of vehicle \(i\) at step \(k\) in proportion to how much the filter changes its next speed and steering angle:
\begin{equation}
r^i_k \leftarrow r^i_k - \zeta_{\mathrm{interv}}\, \Delta t \Big( \tfrac{|a^i_k - a^i_{\mathrm{nom},k}|}{v_{\max}} + \tfrac{|\omega^i_k - \omega^i_{\mathrm{nom},k}|}{\delta_{\max}} \Big),\label{eq:interv}
\end{equation}
where \(\bm u^i_{\mathrm{nom},k} = (a^i_{\mathrm{nom},k}, \omega^i_{\mathrm{nom},k})\) is the saturated nominal control of \cref{sec:problem}, \(\bm u^i_k = (a^i_k, \omega^i_k)\) is the applied one, and the weight \(\zeta_{\mathrm{interv}} > 0\) sets the strength of the penalty. The penalty teaches the policy to propose actions that the filter accepts, so that the fine-tuned policy is safer without the filter and needs fewer interventions with it.

\section{Evaluation}\label{sec:evaluation}

\begin{figure}[t]
\centering
% Overview of the four evaluation stages (column width).
\begin{tikzpicture}[
    font=\scriptsize,
    >=latex,
    stage/.style={draw, rounded corners=1.5pt, align=center, inner sep=2pt, text width=1.22cm,
      minimum height=1.25cm},
    sig/.style={->, thick},
    lbl/.style={font=\tiny, align=center, inner sep=1pt, pos=0.42},
  ]
  \node[stage] (s1) at (0, 0) {\textbf{Stage 1}\\ Pre-training\\ with safety\\ rewards};
  \node[stage] (s2) at (2.4, 0) {\textbf{Stage 2}\\ Zero-shot\\ deployment\\ with filters};
  \node[stage] (s3) at (4.8, 0) {\textbf{Stage 3}\\ Fine-tuning\\ with filters\\ in the loop};
  \node[stage] (s4) at (7.2, 0) {\textbf{Stage 4}\\ Comparison\\ with human\\ drivers};

  \draw[sig] (s1) -- node[lbl, above] {TTC\\ policies\\ (iter.~100)} (s2);
  \draw[sig] (s2) -- node[lbl, above] {same\\ policies} (s3);
  \draw[sig] (s3) -- node[lbl, above] {BH-CBF\\ fine-tuned} (s4);
\end{tikzpicture}
\caption{The evaluation stages along a MARL policy's life cycle.}
\label{fig:eval-stages}
\end{figure}
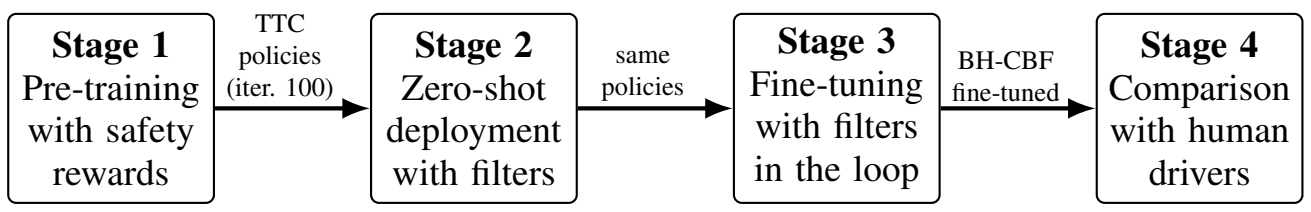

We evaluate our BH-CBF along the life cycle of a MARL policy in four stages outlined in \cref{fig:eval-stages} (arrows: policies passed on to the next stage). Stage~1 in \cref{sec:pretrain}  shows how much safer MARL policies become when pre-trained with our BH-CBF-informed reward instead of a heuristic reward or the one-step CBF reward. Stage~2 in \cref{sec:deploy} deploys unsafe pre-trained policies with different run-time filters zero-shot, i.e., without further training. It shows which filters avoid all collisions, also on unseen maps. Stage~3 in \cref{sec:finetune} shows how much safer fine-tuning with a filter in the loop makes these policies and how many collisions occur during fine-tuning. Stage~4 in \cref{sec:human} compares the fine-tuned, filtered policies with human drivers in recorded traffic. Stages~2 and~3 compare our BH-CBF with state-of-the-art CBF filters and with ablations. \Cref{sec:eval-time} then measures how the computation time of BH-CBF grows with the number of vehicles. We first describe the setup that all stages share in \cref{sec:eval-setup}.

\subsection{Setup}\label{sec:eval-setup}

We simulate in SigmaRL \cite{xu2024sigmarl} with the kinematic bicycle model of \cref{sec:problem}, integrated by explicit Euler with \(\Delta t = \SI{0.1}{\second}\). In Stages~1 to~3, all vehicles have a full-scale footprint of \SI{4.6}{\meter} \(\times\) \SI{2.0}{\meter}, a rear-axle distance \(\ell_{\mathrm r} = \SI{1.35}{\meter}\), and a wheelbase \(\ell_{\mathrm{wb}} = \SI{2.7}{\meter}\). Their accelerations, steering rates, and steering angles are bounded by \(a \in [-6, 3]\,\si{\meter\per\second\squared}\), \(\omega \in [-1, 1]\,\si{\radian\per\second}\), and \(\delta \in [-35, 35]\,\si{\degree}\). Stage~4 uses the actual footprints of the recorded vehicles. The speed limit is \SI{50}{\kilo\meter\per\hour}. We use two maps of drone datasets with recorded human trajectories, both in Aachen, Germany: a roundabout in Neuweiler\footnote{\url{https://www.openstreetmap.org/\#map=19/50.890500/6.174700&layers=V}} from rounD \cite{krajewski2020dataset} (location ID~0) and an unsignalized intersection at Neuk\"ollner Stra\ss e\footnote{\url{https://www.openstreetmap.org/\#map=19/50.785200/6.131100&layers=V}} from inD \cite{bock2020ind} (location ID~4). They require different driving skills: merging into and leaving a ring versus crossing and turning in narrow lanes. We call the map on which a policy was trained its \emph{own map} and the other map its \emph{unseen map}. Stages~1 to~3 use 16 vehicles on the roundabout and 12 on the intersection, and Stage~4 uses the recorded traffic. In Stages~1 to~3, each vehicle follows a route from an entry to an exit of the map, drawn uniformly at random. It spawns at a random point of its route where the joint state satisfies \ref{as:init}. A vehicle reaches its goal when it arrives at the exit of its route and then respawns on a new random route. In Stages~1 and~2, we deploy every policy on its own and its unseen map with three seeds of one simulated minute each.

We use clearances \(\varepsilon_{\mathrm{lane}} = \SI{2}{\centi\meter}\) and \(\varepsilon_{\mathrm{veh}} = \SI{10}{\centi\meter}\). Our exact check accepts an inequality up to a tolerance \(\tau = \SI{0.4}{\milli\meter}\), which lowers the guaranteed bound of \(B_e\) to \(-\tau/\lambda\), or \(-2\tau/\lambda\) for a pair in the decentralized filter. All evaluated filters use \(\lambda \ge 0.1\), so the clearances cover this term (at most \SI{4}{\milli\meter} and \SI{8}{\milli\meter}). In our corner budgets, we set the offset \(\ell_{\mathrm{off}}\) in \eqref{eq:share} to the vehicle length, \(\ell_{\mathrm{off}} = \SI{4.6}{\meter}\). The backup steers toward a centerline point \(\SI{9.2}{\meter} + (\SI{0.2}{\second})\, v^i\) ahead of the vehicle's projection onto the centerline, with a steering rate of \(\SI{4}{\per\second}\) times the steering-angle error. The QP weights the acceleration in \si{\meter\per\second\squared} and the steering rate in \si{\radian\per\second} by \(\bm W^i = \operatorname{diag}(0.1, 1)\) for every vehicle, which the centralized filter stacks into a block-diagonal \(\bm W\). Hence a deviation of \SI{1}{\meter\per\second\squared} costs as much as one of \(\sqrt{0.1}\,\si{\radian\per\second} \approx \SI{0.32}{\radian\per\second}\). Each constraint has a nonnegative slack with a heavy quadratic penalty (weight \(2.5 \times 10^7\)), so the QP uses it only when the linearized inequalities are infeasible.

We report five metrics:
\begin{enumerate*}[label=(\roman*)]
\item collisions: inter-vehicle and lane-boundary collisions per km driven by all vehicles (an inter-vehicle collision counts twice, and a colliding vehicle respawns);
\item throughput: the number of vehicles reaching their goals per second;
\item runs that leave the certified set \(\mathcal C\) in \eqref{eq:certified}: runs with at least one step with some \(B_e < 0\), evaluated with our margins for every filter (the backup prediction from such a step would violate a margin, so the filter has lost its guarantee);
\item fallbacks: the number of vehicles switched to their backup per 1000 agent-steps; and
\item computation time per step: for the fully decentralized filters, a parallelized estimate with one computation unit per vehicle, given by the shared per-step preparation plus the longest time any vehicle needs to solve its filter's optimization problem.
\end{enumerate*}
We time all filters on the same hardware, an Apple M2 Max running ten simulations in parallel on one CPU core each. Unless stated otherwise, we report means over policies. Values after \(\pm\) are standard errors over policies.

\subsection{Stage 1: Pre-Training With Safety Rewards}\label{sec:pretrain}

For each of three safety rewards, we pre-train ten policies (five random seeds on each map) and deploy them zero-shot, both without a filter. We use MAPPO in our SigmaRL framework \cite{xu2024sigmarl} for 500 iterations of 128 steps in 32 parallel environments. The actor and the centralized critic are multilayer perceptrons with three hidden layers of 256 tanh units, and all vehicles share the actor. We use a discount factor of 0.99, generalized advantage estimation with parameter 0.9, a clip ratio of 0.15, an entropy coefficient of 0.005, and five epochs with minibatches of 512 per iteration. The learning rate decays linearly from \(10^{-4}\) to \(10^{-5}\). All policies share the task reward of SigmaRL (progress along the route, a goal reward, and penalties for collisions and deadlocks) and differ in the safety reward:
\begin{enumerate*}[label=(\roman*)]
\item the Time-To-Collision (TTC) reward, a commonly used heuristic \cite{hayward1972nearmiss}, which penalizes a vehicle whose TTC with another vehicle falls below a threshold (computed under constant velocities) and a vehicle that comes close to a lane boundary;
\item the one-step CBF reward of our prior work \cite{xu2026safety}, which has the form of \eqref{eq:cbf-reward} but measures how far the nominal controls violate the one-step condition \eqref{eq:one-step} instead of the backup-horizon condition \eqref{eq:bh-cbf}; and
\item our BH-CBF-informed reward \eqref{eq:cbf-reward}.
\end{enumerate*}
The two CBF rewards share the margins and \(\lambda = 0.2\) and differ only in the condition. We use \(h_{\mathrm{norm},\mathrm L} = h_{\mathrm{norm},\mathrm R} = v_{\max}\Delta t\) for a lane side and \(h_{\mathrm{norm},\mathrm V} = 2 v_{\max}\Delta t\) for a pair as normalizers in \eqref{eq:cbf-reward}, scaling each violation by the most a margin can shrink in one step.

\begin{figure}[t]
\centering
\includegraphics[width=\columnwidth]{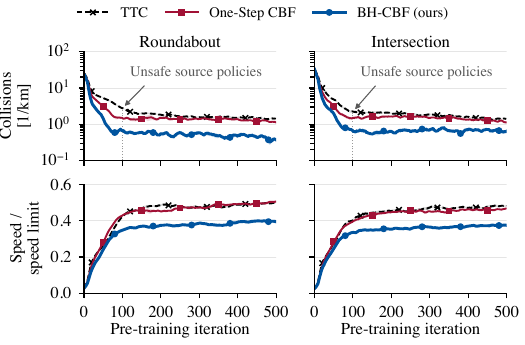}
\caption{Stage 1: training rollouts over pre-training by safety reward, smoothed over ten iterations, mean and standard error over five training seeds.}
\label{fig:pretrain}
\end{figure}

\Cref{fig:pretrain} shows the training rollouts over pre-training: our BH-CBF-informed reward lowers collisions rapidly within the first 80 iterations and keeps them low (blue in the top row). At iteration 500, its policies collide 0.38--0.64 times per km on the two maps, against 1.12--1.41 with the other rewards. They drive at \SIrange{38}{40}{\percent} of the speed limit, whereas the policies of the other rewards drive at \SIrange{47}{51}{\percent} (bottom row).

\Cref{tab:pretrain-reward} shows the pre-trained policies (at iteration 500) deployed without a filter: our BH-CBF-informed reward trains the safest policies. Unlike in training, the deployed policies no longer vary their actions randomly to explore new behavior, which lowers their collisions. On their own map, they collide \(0.05 \pm 0.02\) times per km, against \(0.86 \pm 0.12\) and \(0.71 \pm 0.06\) for the TTC and the one-step CBF reward, at a similar throughput. The one-step CBF reward hardly improves on the TTC heuristic, so the gain comes from our backup horizon. On the unseen map, every policy collides, also ours. Therefore, learning alone does not transfer safety to new maps, which motivates the run-time filter.

\begin{table}[t]
\centering
\footnotesize
\caption{Stage 1: the 30 pre-trained policies (ten per safety reward, iteration 500) deployed without a filter.}
\label{tab:pretrain-reward}
\setlength{\tabcolsep}{4.5pt}
\begin{tabular}{@{}lllll@{}}
\toprule
& \multicolumn{2}{c}{Own map} & \multicolumn{2}{c}{Unseen map} \\
\cmidrule(lr){2-3}\cmidrule(l){4-5}
& Collisions\,\(\downarrow\) & Throughput\,\(\uparrow\) & Collisions\,\(\downarrow\) & Throughput\,\(\uparrow\) \\
Reward method & [1/km] & [veh/s] & [1/km] & [veh/s] \\
\midrule
TTC & 0.86 & 1.08 & 1.27 & 0.92 \\
One-Step CBF & 0.71 & 1.03 & 1.61 & 0.80 \\
\textbf{BH-CBF (ours)} & \textbf{0.05} & 1.05 & \textbf{0.59} & 0.91 \\
\bottomrule
\end{tabular}
\end{table}

\subsection{Stage 2: Zero-Shot Deployment With Run-Time Filters}\label{sec:deploy}

\begin{table}[t]
\centering
\footnotesize
\caption{Compared safety filters; gray: as in BH-CBF.}
\label{tab:eval-filters}
\setlength{\tabcolsep}{4pt}
\begin{tabular}{@{}llll@{}}
\toprule
Filter & Geometry & Condition & Split \\
\midrule
\textbf{BH-CBF (ours)} & exact & backup horizon & corner budgets \\
\noalign{\vskip 1pt}\hdashline\noalign{\vskip 2pt}
A-Geometry & capsules & \asours{backup horizon} & \asours{corner budgets} \\
A-Condition & \asours{exact} & one step & \asours{corner budgets} \\
A-Split & \asours{exact} & \asours{backup horizon} & half split \\
A-Centralized & \asours{exact} & \asours{backup horizon} & centralized \\
\noalign{\vskip 1pt}\hdashline\noalign{\vskip 2pt}
B-SAT \cite{wu2025optimizationfree} & smoothed SAT & one step & half split \\
B-Frenet \cite{voros2025lanekeeping,squires2025barrier} & Frenet lanes & one step & our shares \\
B-Disk \cite{wang2017safety} & disk pairs & one step & half split \\
B-Dual \cite{thirugnanam2022dualitybased,thirugnanam2022safetycritical} & exact (convex) & one step & half split \\
B-Backup \cite{chen2021backup,chen2021guaranteed} & \asours{exact} & fixed horizon & half split \\
B-MPS \cite{bastani2021safea,zhang2019mamps} & \asours{exact} & accept or fall back & -- \\
\bottomrule
\end{tabular}
\end{table}

\begin{table}[t]
\centering
\footnotesize
\caption{Stage 2: the ten unsafe source policies (TTC reward, iteration 100) deployed with fully decentralized filters, except A-Centralized; 60 runs (own and unseen map, three seeds each) per row. Runs \(\notin \mathcal C\): runs that leave the certified set; time: computation time per step.}
\label{tab:deploy-filters}
\setlength{\tabcolsep}{3.5pt}
\begin{tabular}{@{}llllll@{}}
\toprule
& Collisions\,\(\downarrow\) & Runs\,\(\downarrow\) & Fallbacks\,\(\downarrow\) & Throughput\,\(\uparrow\) & Time\,\(\downarrow\) \\
Filter & [1/km] & \(\notin \mathcal C\) & [1/1000] & [veh/s] & [ms] \\
\midrule
Unfiltered & 1.39 & 60/60 & -- & 0.87 & -- \\
\noalign{\vskip 1pt}\hdashline\noalign{\vskip 2pt}
B-Frenet & 1.22 & 60/60 & -- & 0.75 & 4.7 \\
A-Condition & 0.61 & 60/60 & -- & 0.91 & 8.0 \\
B-Disk & 0.47 & 60/60 & -- & 1.04 & 7.1 \\
B-SAT & 0.38 & 60/60 & -- & 0.91 & 11.3 \\
B-Dual & 0.25 & 60/60 & -- & 1.12 & 15.8 \\
\noalign{\vskip 1pt}\hdashline\noalign{\vskip 2pt}
B-Backup & 0 & 14/60 & 3.3 & 1.19 & 15.5 \\
A-Split & 0 & 4/60 & 4.7 & 1.21 & 14.2 \\
B-MPS & 0 & 0/60 & 65.7 & 1.18 & 12.3 \\
A-Centralized & 0 & 0/60 & 7.4 & 1.21 & 80.6 \\
A-Geometry & 0 & 0/60 & 6.9 & 1.20 & 9.0 \\
\textbf{BH-CBF (ours)} & \textbf{0} & \textbf{0/60} & \textbf{3.3} & 1.20 & 12.6 \\
\bottomrule
\end{tabular}
\end{table}

\begin{figure*}[!t]
\centering
\begin{subfigure}[t]{0.119\textwidth}
\centering
\includegraphics[width=\linewidth]{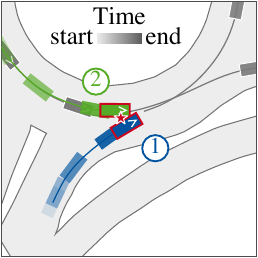}\\[1pt]
\includegraphics[width=\linewidth]{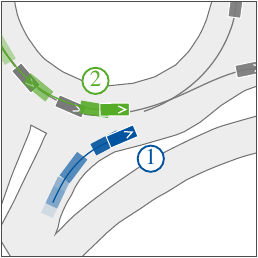}
\caption{}
\label{fig:cf-a}
\end{subfigure}\hfill
\begin{subfigure}[t]{0.119\textwidth}
\centering
\includegraphics[width=\linewidth]{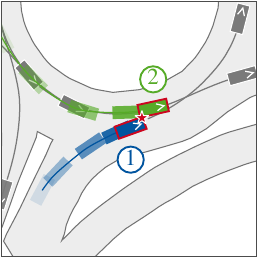}\\[1pt]
\includegraphics[width=\linewidth]{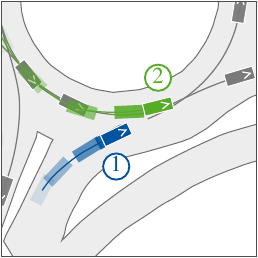}
\caption{}
\label{fig:cf-b}
\end{subfigure}\hfill
\begin{subfigure}[t]{0.119\textwidth}
\centering
\includegraphics[width=\linewidth]{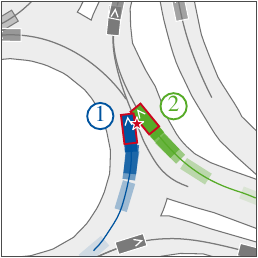}\\[1pt]
\includegraphics[width=\linewidth]{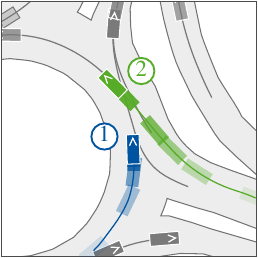}
\caption{}
\label{fig:cf-c}
\end{subfigure}\hfill
\begin{subfigure}[t]{0.119\textwidth}
\centering
\includegraphics[width=\linewidth]{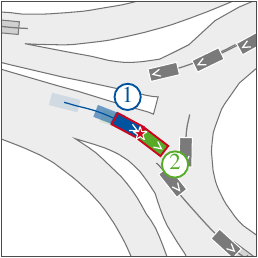}\\[1pt]
\includegraphics[width=\linewidth]{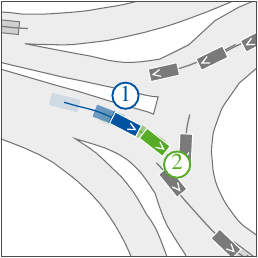}
\caption{}
\label{fig:cf-d}
\end{subfigure}\hspace{0.012\textwidth}\hfill
\begin{subfigure}[t]{0.119\textwidth}
\centering
\includegraphics[width=\linewidth]{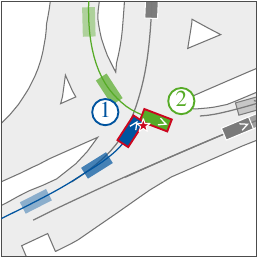}\\[1pt]
\includegraphics[width=\linewidth]{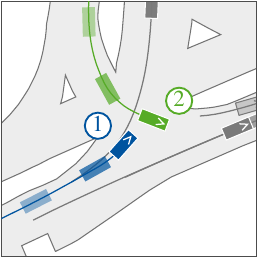}
\caption{}
\label{fig:cf-e}
\end{subfigure}\hfill
\begin{subfigure}[t]{0.119\textwidth}
\centering
\includegraphics[width=\linewidth]{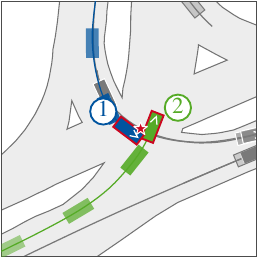}\\[1pt]
\includegraphics[width=\linewidth]{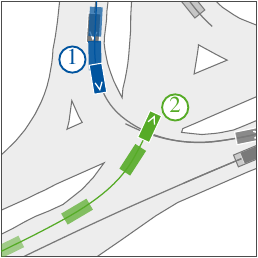}
\caption{}
\label{fig:cf-f}
\end{subfigure}\hfill
\begin{subfigure}[t]{0.119\textwidth}
\centering
\includegraphics[width=\linewidth]{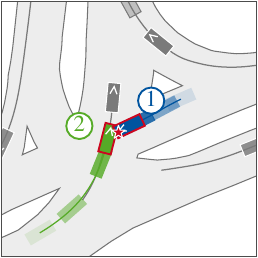}\\[1pt]
\includegraphics[width=\linewidth]{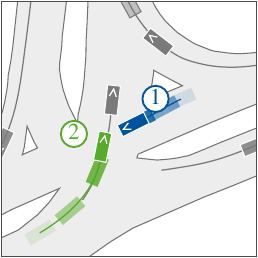}
\caption{}
\label{fig:cf-g}
\end{subfigure}\hfill
\begin{subfigure}[t]{0.119\textwidth}
\centering
\includegraphics[width=\linewidth]{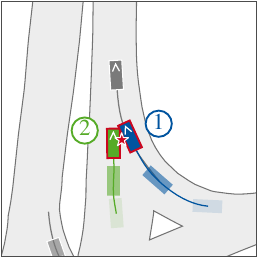}\\[1pt]
\includegraphics[width=\linewidth]{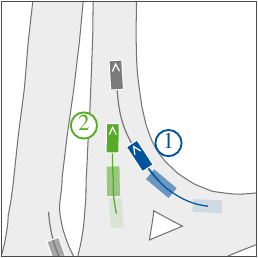}
\caption{}
\label{fig:cf-h}
\end{subfigure}
\caption{Unsafe policies on the roundabout (a)--(d) and the intersection (e)--(h), without (top) and with (bottom) our BH-CBF filter. Blue: yielding vehicle~1; green: vehicle~2; red: collisions; gray: others (last footprint only); footprint interval: \SI{1.0}{\second}.}
\label{fig:footprints}
\end{figure*}

We deploy unsafe pre-trained policies zero-shot with our decentralized BH-CBF filter, with ablations of our design, and with CBF filters from the literature. To challenge the filters, we use the ten TTC-reward policies of pre-training iteration~100 as source policies. They drive fast and collide often (arrows to the dashed black lines in \cref{fig:pretrain}: 2.2--2.8 collisions per km on the two maps). Stage~3 fine-tunes the same source policies.

For a fair comparison, all filters share the nominal controls, the QP solver, and the fallback to the backup. \Cref{tab:eval-filters} shows how they differ: only in their margins (column Geometry), their CBF condition, and their decentralized split. All filters except A-Centralized run fully decentralized. The prefix ``A'' marks an ablation and ``B'' a baseline. Each of our four ablations changes one of these components of BH-CBF:
\begin{enumerate*}[label=(\roman*)]
\item A-Geometry replaces our exact vehicle geometry with a tight convex capsule and applies the corner budgets to the end points of the capsule axis;
\item A-Condition replaces our backup-horizon condition \eqref{eq:bh-cbf} with the standard one-step CBF condition \eqref{eq:one-step};
\item A-Split replaces our corner budgets \eqref{eq:dec-corner} with the half split of the decentralized backup CBF \cite{chen2021guaranteed}, in which each vehicle keeps its candidate prediction away from the backup prediction of the other and covers half of the allowed decrease of \(B^{ij}\); and
\item A-Centralized solves the centralized filter problem \eqref{eq:filter} for all vehicles jointly.
\end{enumerate*}
The first four baselines use the one-step condition \eqref{eq:one-step} with the barrier functions of state-of-the-art CBF filters instead of our margins. The last two are backup-based safeguards with our margins:
\begin{enumerate*}[label=(\roman*)]
\item B-SAT \cite{wu2025optimizationfree}, named after the Separating Axis Theorem (SAT), replaces both the lane and the pair inequalities and splits each pair in half;
\item B-Frenet \cite{voros2025lanekeeping,squires2025barrier} replaces only the lane inequalities, keeps our pair margins, and splits each pair with our shares \eqref{eq:share};
\item B-Disk \cite{wang2017safety} replaces only the pair inequalities, keeps our lane inequalities, and splits each pair in half;
\item B-Dual \cite{thirugnanam2022dualitybased,thirugnanam2022safetycritical} replaces both the lane and the pair inequalities and uses the same half split;
\item B-Backup is the finite-step backup CBF \cite{chen2021backup,chen2021guaranteed} with a fixed horizon and its half split; and
\item B-MPS is model predictive shielding \cite{bastani2021safea,zhang2019mamps} with our backup, which replaces a nominal control that fails the check of \cref{rem:backup-cbf} by the backup.
\end{enumerate*}
We tune the decay rate \(\lambda\) and the method-specific parameters of every filter, such as the smoothing parameter of B-SAT. Among the parameter values that are nearly as safe as the safest, we choose those with the best throughput.

\Cref{tab:deploy-filters} shows the source policies deployed with each decentralized filter: only the backup-based filters avoid all collisions, both on their own and on the unseen map. The one-step filters collide mostly at the lane boundaries (B-Dual mostly between vehicles). Every one of these collisions follows a step with an infeasible QP, and these filters need slack in \SIrange{18}{59}{\percent} of the steps. The ablations show that the one-step condition causes these collisions, not the approximated vehicle geometry: A-Condition collides although it uses our exact margins, whereas A-Geometry is collision-free although it approximates the vehicle geometry. Without collisions, more vehicles reach their exits, which raises the throughput from 0.87 to 1.18--1.21\,veh/s.
Among the collision-free filters, our BH-CBF is the only one that keeps the guarantee with few fallbacks and without overapproximating the vehicle geometry. B-Backup and A-Split leave the certified set in 14 and 4 of 60 runs (smallest backup values down to \SI{-1.21}{\meter}). In these runs, the backup loses its safety guarantee, although no collision followed. A-Split differs from BH-CBF only in how the pair condition is split between the two vehicles. Its half split bounds each vehicle against the backup prediction of the other, which does not bound their distance when both deviate. Our corner budgets bound both vehicles along a common separating normal, so their deviations add up to at most the budget for any complementary shares (\cref{thm:decentralized}). A centralized filter brings no advantage: A-Centralized is equally safe and reaches the same throughput, but it falls back more often (7.4 against 3.3 times per 1000 agent-steps). When a pair inequality fails the exact check, its fallback switches both vehicles of the pair to their backup, whereas in our decentralized filter each vehicle switches only itself (\cref{sec:filter}). B-MPS stays certified by falling back 65.7 times per 1000 agent-steps, against 3.3 for BH-CBF. A-Geometry also stays certified but overapproximates the vehicle geometry. At the same throughput, it thus falls back twice as often. The decentralized BH-CBF needs \SI{12.6}{\milli\second} per step, far below the sampling period \(\Delta t=\SI{100}{\milli\second}\). The decay rate regulates how early our BH-CBF intervenes, but safety does not hinge on it: with \(\lambda \in \{0.1, 0.2, 0.5, 1\}\), it stays collision-free and certified, and only \(\lambda = 0.1\) lowers the throughput (1.02 against 1.20\,veh/s for the tuned \(\lambda = 0.5\)).

\Cref{fig:footprints} shows representative collisions of the unsafe source policies. In each panel, the top row shows the unfiltered policies and the bottom row the policies filtered by our BH-CBF, both from the same initial joint state. Without the filter, the vehicles collide (red). With the filter, one vehicle yields by braking or by merging into a gap (blue vehicle~1), and no collision occurs.

\begin{figure}[t]
\centering
\begin{subfigure}{\columnwidth}
\centering
\includegraphics[width=\linewidth]{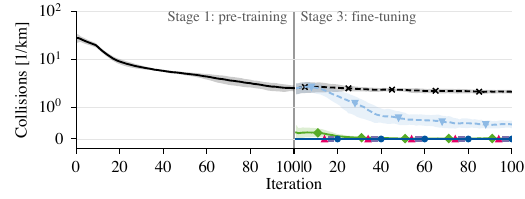}
\caption{Training rollouts, smoothed over ten iterations.}
\label{fig:finetune-train}
\end{subfigure}
\begin{minipage}[t]{0.5\columnwidth}
\vspace{0pt}
\includegraphics[width=\linewidth]{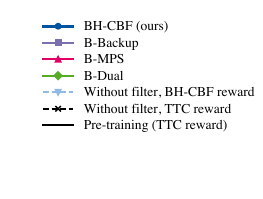}
\end{minipage}\hfill
\begin{subfigure}[t]{0.5\columnwidth}
\vspace{0pt}
\centering
\includegraphics[width=\linewidth]{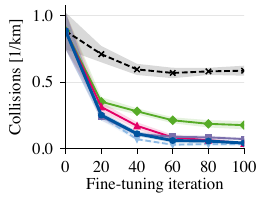}
\caption{Deployed without a filter.}
\label{fig:finetune-eval}
\end{subfigure}
\caption{Collisions of the TTC-reward policies over pre-training and fine-tuning, mean over runs (20 fine-tuning runs per setting, own and unseen map taken together) with standard deviation in (\subref{fig:finetune-train}) and standard error in (\subref{fig:finetune-eval}).}
\label{fig:finetune}
\end{figure}

\subsection{Stage 3: Fine-Tuning With Filters in the Loop}\label{sec:finetune}

A policy that collides on a new map should improve there, but collisions during training are not acceptable in real-world operation. Therefore, we fine-tune the unsafe source policies of Stage~2 with a filter in the loop and the intervention penalty \eqref{eq:interv} with \(\zeta_{\mathrm{interv}} = 3\). Each policy is fine-tuned on its own and on its unseen map for 100 iterations of 128 steps in 16 parallel environments. We compare our BH-CBF with the three fully decentralized baselines that collided least in Stage~2: B-Backup, B-MPS, and B-Dual. As ablations, we fine-tune without a filter, once with the TTC reward of the pre-training and once with our BH-CBF-informed reward. Each setting has 20 runs (ten source policies, two target maps). During fine-tuning, we deploy the policies every 20 iterations (from iteration 0) for eight simulated minutes without a filter and with the filter of their fine-tuning.

\Cref{fig:finetune-train} shows the collisions of the training rollouts: only the backup-based filters keep fine-tuning collision-free in all 20 runs (BH-CBF, B-Backup, and B-MPS in blue, purple, and magenta at zero after iteration 100). Without a filter, the policies collide on average over fine-tuning 2.29 times per km with the TTC reward (dashed black line) and 0.96 times per km with our BH-CBF-informed reward, which teaches them to avoid collisions only gradually (dashed light blue line). B-Dual collides 0.05 times per km (green line). However, B-Backup leaves \(\mathcal C\) in 19 of its 20 runs, since it does not certify the joint backup.

\Cref{fig:finetune-eval} shows the fine-tuned policies deployed without a filter: fine-tuning with BH-CBF in the loop makes the policy itself safe (blue line). Its collisions drop by \SI{95}{\percent}, from \(0.89 \pm 0.14\) to \(0.04 \pm 0.01\) per km, most of it within 40 iterations. Fine-tuning without a filter and with the TTC reward only reaches \(0.59 \pm 0.04\) (dashed black line). The intervention penalty thus turns the filter's corrections into a learning signal that lowers collisions as much as our BH-CBF-informed reward (dashed light blue line), but without the collisions that this reward needs during training. The other filters end at 0.04 (B-MPS), 0.07 (B-Backup), and 0.18 (B-Dual). The policy does not become safe by stopping: its throughput rises from 0.86 to 1.02\,veh/s. Deployed with the BH-CBF filter, the fine-tuned policy also needs fewer interventions than the source policy: the filter changes \SI{6.1}{\percent} of its actions instead of \SI{13.7}{\percent}. Fallbacks become rare as well: 0.23 instead of 3.1 per 1000 agent-steps.

\subsection{Stage 4: Comparison With Human Drivers}\label{sec:human}

Stage~4 compares our BH-CBF with human drivers. We take the ten policies of Stage~3 that are fine-tuned with our BH-CBF filter on the roundabout: five pre-trained on the roundabout and five on the intersection. We deploy them with our decentralized BH-CBF filter on the roundabout and compare them with the human drivers recorded on the same roundabout in rounD \cite{krajewski2020dataset}. We extract the time windows of rounD in which only cars and vans are present. This excludes road users that we cannot control, such as pedestrians. The 112 windows contain 1264 vehicles (1213 cars and 51 vans), which drive \SI{115}{\kilo\meter} in 4.4 vehicle-hours in total.

Every human-driven vehicle of a window is replaced by a controlled vehicle that enters at the recorded time and state and follows the human's route to the human's last recorded position. Unlike in Stages~1 to~3, the vehicles are heterogeneous, and the filter certifies safety under individual vehicle geometries and acceleration bounds. Each vehicle has its recorded footprint. Since the dataset provides no control bounds or axle positions, we use the acceleration bounds \([-6, 3]\,\si{\meter\per\second\squared}\) for cars and \([-5, 2.5]\,\si{\meter\per\second\squared}\) for vans. We scale the rear-axle distance and the wheelbase with the vehicle length, using the ratios of the nominal \SI{4.6}{\meter} vehicle of \cref{sec:eval-setup}.

\begin{table}[t]
\centering
\footnotesize
\caption{Stage 4: Human-comparison results over 115 km of 1264 recorded human-driven vehicles. Stops: stop before the ring entry; gap: 5th percentile of the minimum gap.}
\label{tab:human}
\setlength{\tabcolsep}{4.5pt}
\begin{tabular}{@{}lllll@{}}
\toprule
& Travel time\,\(\downarrow\) & Stops\,\(\downarrow\) & Gap & RMS jerk\,\(\downarrow\) \\
Controlled by & median [s] & [\%] & [m] & median [\si{\meter\per\second\cubed}] \\
\midrule
Humans & 12.0 & 13.8 & 1.40 & 0.94 \\
\textbf{BH-CBF (ours)} & \textbf{10.3} & \textbf{2.0} & 1.22 & 3.66 \\
\bottomrule
\end{tabular}
\end{table}

\begin{figure}[t]
\centering
\begin{subfigure}[t]{0.49\columnwidth}
\centering
\includegraphics[width=\linewidth]{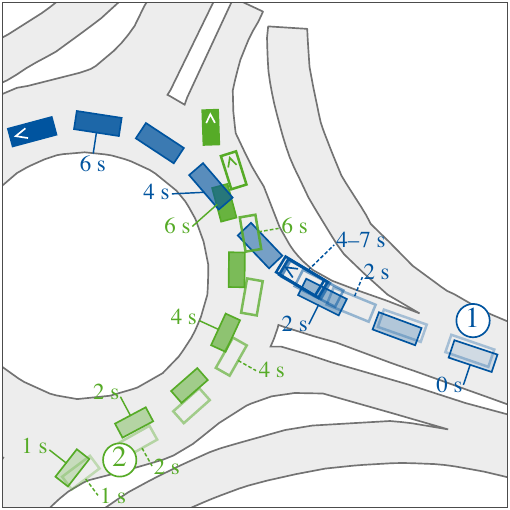}
\caption{}
\label{fig:human-keep-speed}
\end{subfigure}\hfill
\begin{subfigure}[t]{0.49\columnwidth}
\centering
\includegraphics[width=\linewidth]{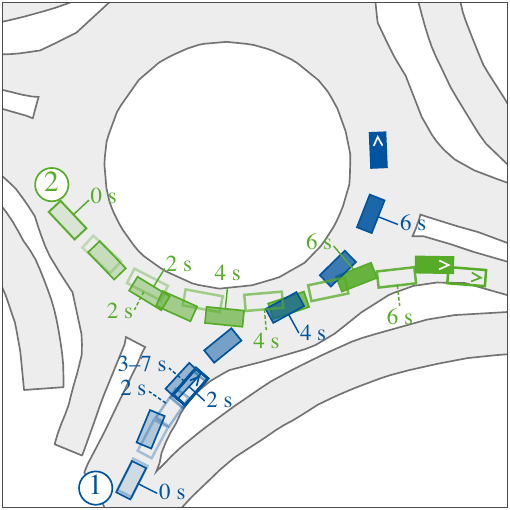}
\caption{}
\label{fig:human-close-gap}
\end{subfigure}
\caption{Stage 4: two roundabout scenarios over \SI{7}{\second}. Filled: our BH-CBF-filtered policies; unfilled: recorded humans (time stamps with dashed lines). Other vehicles not shown.}
\label{fig:human-footprints}
\end{figure}

\Cref{tab:human} presents the comparison results. With our BH-CBF-filtered policies, no vehicle collides or touches a lane boundary, every vehicle reaches its goal, and the joint state stays in \(\mathcal C\) at every step. Overall, our policies reduce the median travel time by \SI{14}{\percent}, from \SI{12.0}{\second} to \SI{10.3}{\second}. The policies save this time mainly at the entry of the ring: they stop before the entry in \SI{2.0}{\percent} of the cases, against \SI{13.8}{\percent} for the humans. They also keep smaller minimum gaps between vehicles than the humans and thus use the road space more efficiently. One drawback of the policies is lower comfort: their median Root-Mean-Square (RMS) jerk is about four times that of the humans, since neither the reward nor the filter penalizes jerk. The filter needs about \SI{14}{\milli\second} per step, far below the sampling period \(\Delta t = \SI{100}{\milli\second}\).

\Cref{fig:human-footprints} shows two typical scenarios in which the human (unfilled footprints) stops and the BH-CBF-filtered policy (filled footprints) does not. In \cref{fig:human-keep-speed}, vehicle~1 in blue approaches at \SI{10}{\meter\per\second} and keeps its speed into the ring ahead of vehicle~2. The human instead brakes to a standstill and waits \SI{3}{\second} for vehicle~2. In \cref{fig:human-close-gap}, the policy enters ahead of a close circulating vehicle~2. Vehicle~2 then brakes briefly from \SI{8.8}{\meter\per\second} to about \SI{3}{\meter\per\second} and loses a small amount of time, whereas the human driver of vehicle~1 waits \SI{5.4}{\second} for vehicle~2 and two further circulating vehicles (not shown). The policy can take this gap because the filter does not encode the right of way that humans follow and admits every entry that it certifies as safe. In this scenario, a short delay of the circulating vehicle spares the entering vehicle a much longer wait.

\begin{figure}[t]
\centering
\begin{subfigure}[b]{0.40\columnwidth}
\centering
\includegraphics[width=\linewidth]{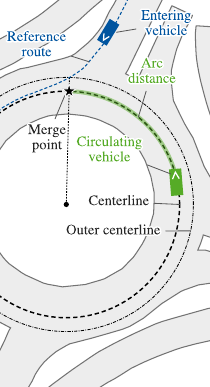}
\caption{Merge point (schematic).}
\label{fig:gap-scenario}
\end{subfigure}\hfill
\begin{minipage}[b]{0.58\columnwidth}
\begin{subfigure}{\linewidth}
\centering
\includegraphics[width=\linewidth]{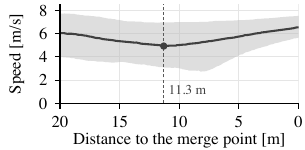}
\caption{Approach speed of the humans.}
\label{fig:gap-speed}
\end{subfigure}\\[2pt]
\begin{subfigure}{\linewidth}
\centering
\includegraphics[width=\linewidth]{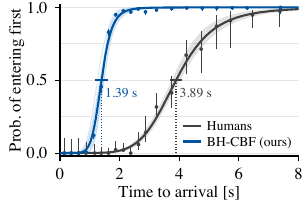}
\caption{Gap acceptance.}
\label{fig:gap-probability}
\end{subfigure}
\end{minipage}
\caption{Stage 4: entering the ring. (b)~Median and \SIrange{25}{75}{\percent} band. (c)~Probability of entering the ring before the circulating vehicle (dots: share per time-to-arrival bin with \SI{95}{\percent} confidence intervals; thick curves: logistic fits with \SI{95}{\percent} confidence bands; thin blue curves: logistic fits of the ten single policies).}
\label{fig:gap-acceptance}
\end{figure}

To further analyze how the BH-CBF-filtered policies and the humans enter the ring, we estimate the probability that an entering vehicle enters the ring before the next circulating vehicle. We use the \SI{115}{\kilo\meter} of driving data of both the policies and the humans. This probability depends on the time to arrival, which is the time that the circulating vehicle needs to reach the merge point of the entering vehicle. We construct this merge point in three steps, as depicted in \cref{fig:gap-scenario}. First, we take the outer centerline (dash-dotted) and the centerline (dashed) of the ring. Second, the point where the reference route of the entering vehicle crosses the outer centerline defines a ray from the ring center (dotted). Third, the merge point is the intersection of this ray with the centerline (star). The time to arrival is the arc distance of the circulating vehicle to the merge point along the centerline (green dashed) divided by its speed averaged over the last second. It remains to determine the distance to the merge point at which the entering vehicle judges the time to arrival. We derive this distance from the approach speed of the humans shown in \cref{fig:gap-speed}, since humans slow down while they judge the gap and speed up once they have decided to enter. Their median speed is lowest \SI{11.3}{\meter} before the merge point (dot), so we evaluate the time to arrival at this distance. \Cref{fig:gap-probability} shows the probability of entering first over the time to arrival. We fit this probability by a logistic regression in the logarithm of the time to arrival. The fit reaches 0.5 at \SI{3.89}{\second} for the humans (black) and at \SI{1.39}{\second} for the policies (blue). Hence, the policies are likely to enter ahead of a circulating vehicle that arrives in more than \SI{1.39}{\second}, whereas the humans give way to it until its time to arrival exceeds \SI{3.89}{\second}. The policies enter whenever the BH-CBF filter certifies safety instead of waiting, even if the circulating vehicle has to brake (as in \cref{fig:human-close-gap}). This reduces the overall travel time by \SI{14}{\percent} as shown in~\cref{tab:human}. Among all vehicles, from their initial position to the merge point, the policies need \SI{2.59}{\second} on average instead of the humans' \SI{4.50}{\second}, whereas the time from the merge point to the goal remains similar (\SI{8.80}{\second} against \SI{9.04}{\second}). Hence, about \SI{90}{\percent} of the saved time is gained before the merge point. Connectivity contributes to this gain: each CAV knows the states and routes of the circulating vehicles (\ref{as:coop}) and can thus certify a small gap as safe. Without knowing their intentions, a human driver instead waits for a large time gap so as not to hinder the circulating vehicles, which have the right of way.

\subsection{Computation Time}\label{sec:eval-time}

\begin{figure}[t]
\centering
\includegraphics[width=\columnwidth]{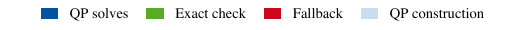}\\[1pt]
\begin{subfigure}[t]{0.5\columnwidth}
\centering
\includegraphics[width=\linewidth]{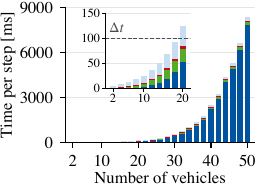}
\caption{Centralized.}
\label{fig:time-cen}
\end{subfigure}\hfill
\begin{subfigure}[t]{0.5\columnwidth}
\centering
\includegraphics[width=\linewidth]{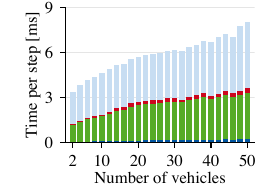}
\caption{Decentralized.}
\label{fig:time-dec}
\end{subfigure}
\caption{Computation time of the BH-CBF filter per step on the roundabout, mean over 500 steps per bar; sampling period \(\Delta t = \SI{100}{\milli\second}\).}
\label{fig:time}
\end{figure}

To determine how the computation time of our decentralized BH-CBF filter grows with the number of vehicles, we measure it on the roundabout with 2 to 50 vehicles, from free flow to congested traffic. For comparison, we also measure the centralized BH-CBF filter. The nominal controls come from the five unsafe TTC-reward source policies of Stage~1 that were pre-trained on the roundabout, which the filter must correct often. For each vehicle count, we run each of the five policies for 100 steps, which gives 500 measured steps per vehicle count.

\Cref{fig:time} shows the computation time per step: the decentralized filter scales well to dense traffic, whereas the centralized filter does not. Each bar splits the time into the QP solves, the exact check, the fallback, and the construction of the QP, i.e., the backup predictions, the linearization of the QP inequalities, and the assembly of the constraint rows. The centralized filter solves one QP for all vehicles, and its time grows superlinearly as in \cref{fig:time-cen}. From 20 vehicles on, it exceeds the sampling period \(\Delta t\). \Cref{fig:time-dec} depicts the computation time of our decentralized filter, which solves one small QP per vehicle with the neighbors within its reach: the mean computation time grows only slowly, from \SI{3.3}{\milli\second} to \SI{8.0}{\milli\second}. Its 95th percentile saturates: from 18 vehicles on, it stays between \SI{12}{\milli\second} and \SI{13.8}{\milli\second}, far below the sampling period \(\Delta t\). Nearly all of this time is the QP construction and the exact check, which grow only with the number of neighbors. This number saturates, because vehicles in dense traffic drive slower and thus reach less far: from 30 to 50 vehicles, the most constrained vehicle has 9 to 12 neighbors. The QP solves take at most \SI{0.2}{\milli\second} and the fallback at most \SI{0.4}{\milli\second}. Decentralization also costs little performance in dense traffic. Both filters avoid all collisions at every vehicle count, and the mean speed under our decentralized filter stays within \SI{1}{\percent} of the centralized one up to 20 vehicles and at most \SI{5}{\percent} below it beyond.

\section{Conclusion}\label{sec:conclusion}

We proposed BH-CBF, a recursively feasible CBF for CAVs that considers the exact vehicle and road geometry, and a fully decentralized safety filter based on it. Our results lead to four conclusions. First, a CBF filter must be recursively feasible, since recursive feasibility decides whether it can avoid collisions with a strict guarantee. Second, one barrier function can safeguard the whole learning life cycle. As a reward, BH-CBF trains policies that are nearly collision-free on their own map; as a run-time filter, it guarantees safety for any policy, so learning can focus on efficiency; during fine-tuning with the filter in the loop, its corrections become a learning signal that makes the policy itself safe, without a single collision during training. Third, centralization brings no benefit in our experiments. A centralized BH-CBF filter is as safe as our decentralized filter and reaches the same throughput, but it is no longer real-time from 20 vehicles on. Our decentralized filter stays real-time as the number of vehicles grows, since it needs a single broadcast per step and its computation time depends mainly on the number of neighbors. Fourth, connectivity makes safe driving also efficient. Since each vehicle knows the states and routes of its neighbors, our decentralized BH-CBF filter can certify small gaps as safe that human drivers leave unused. In recorded traffic of a roundabout in Germany, our BH-CBF-filtered policies shortened the median travel time by \SI{14}{\percent} compared with the human drivers, without any collision.

The comparison with human drivers covers only part of real traffic. The recorded windows contain cars and vans but no vulnerable road users, such as pedestrians and cyclists, who do not cooperate with the filter. All vehicles are assumed to be CAVs. Moreover, the filtered motion is jerkier than human driving. Future work will address vulnerable road users, mixed traffic with human drivers, and the comfort of the filtered motion, and will validate the method experimentally in our lab.

\section*{Acknowledgment}
Computations were performed with computing resources granted by RWTH Aachen University under project rwth1763.

\bibliographystyle{IEEEtran}
\bibliography{00_literature}

\end{document}